\documentclass[a4paper,fleqn]{cas-dc}%

\usepackage[numbers]{natbib}
\usepackage{graphicx} 
\usepackage{subcaption}
\usepackage{xcolor,listings}
\usepackage{amsmath}
\usepackage{algorithm}
\usepackage{algpseudocode}
\usepackage{array}
\usepackage{pdflscape}
\usepackage{rotating}
\usepackage{svg}
\usepackage{pgffor} 
\usepackage{physics}
\usepackage{tikz}
\usepackage{mathdots}
\usepackage{cancel}
\usepackage{siunitx}
\usepackage{multirow}
\usepackage{amssymb}
\usepackage{gensymb}
\usepackage{tabularx}
\usepackage{extarrows}
\usepackage{booktabs}
\usepackage{tikz}
\usepackage{enumitem}
\usepackage{nicematrix}
\usetikzlibrary{fadings,patterns,shadows.blur,shapes}
\usepackage{overpic}

\def\tsc#1{\csdef{#1}{\textsc{\lowercase{#1}}\xspace}}
\tsc{WGM}
\tsc{QE}

\begin{document}
\let\WriteBookmarks\relax
\def\floatpagepagefraction{1}
\def\textpagefraction{.001}

\shorttitle{Clustering and analysis of user behaviour in blockchain: A case study of Planet IX}    

\shortauthors{Zelenyanszki et al.}  

\title [mode = title]{Clustering and analysis of user behaviour in blockchain: A case study of Planet IX}   



%

\author[1]{Dorottya Zelenyanszki}[orcid=0009-0004-1300-0357]

\cormark[1]


\ead{dora.zelenyanszki@griffithuni.edu.au}



\affiliation[1]{organization={Griffith University},
            country={Australia}}

\author[1]{Zh\'e H\'ou}[orcid=0000-0001-7164-0580]


\ead{z.hou@griffith.edu.au}





\author[2]{Kamanashis Biswas}[orcid=0000-0003-3719-8607]


\ead{kamanashis.biswas@acu.edu.au}

\affiliation[2]{organization={Australian Catholic University},
            country={Australia}}

\author[1]{Vallipuram Muthukkumarasamy}[orcid=0000-0002-6787-6379]


\ead{v.muthu@griffith.edu.au}



\begin{abstract}
Decentralised applications (dApps) that run on public blockchains have the benefit of trustworthiness and transparency as every activity that happens on the blockchain can be publicly traced through the transaction data. However, this introduces a potential privacy problem as this data can be tracked and analysed, which can reveal user-behaviour information. A user behaviour analysis pipeline was proposed to present how this type of information can be extracted and analysed to identify separate behavioural clusters that can describe how users behave in the game. The pipeline starts with the collection of transaction data, involving smart contracts,
that is collected from a blockchain-based game called Planet IX. Both the raw transaction information and the transaction events are considered in the data collection. From this data, separate game actions can be formed and those are leveraged to present how and when the users conducted their in-game activities in the form of user flows. An extended version of these user flows also presents how the Non-Fungible Tokens (NFTs) are being leveraged in the user actions. The latter is given as input for a Graph Neural Network (GNN) model to provide graph embeddings for these flows which then can be leveraged by clustering algorithms to cluster user behaviours into separate behavioural clusters. We benchmark and compare well-known clustering algorithms as a part of the proposed method. The user behaviour clusters were analysed and visualised in a graph format. It was found that behavioural information can be extracted regarding the users that belong to these clusters. Such information can be exploited by malicious users to their advantage. To demonstrate this, a privacy threat model was also presented based on the results that correspond to multiple potentially affected areas.
\end{abstract}





\maketitle

\section{Introduction}
Public blockchains have the benefits of immutability, decentralisation and anonymity \cite{DEEPA2022209}, and have a wide range of potential use cases, such as distributed identity management~\cite{YANG2020102050}, record linkage~\cite{NOBREGA2021101826} and metaverse~\cite{HUYNHTHE2023401}. DApps that run on top of blockchains and NFTs that represent various types of physical and digital objects, such as arts~\cite{10.1145/3474355}, have become widely used in recent years. However, every submitted blockchain transaction can be publicly traced by anyone, thus, the transaction data can be analysed, which can uncover behavioural information. This is somewhat analogous to being able to analyse people's browser history and activities, which is a serious privacy concern. Zhang et al.~\cite{10.1145/3316481} described this as the issue of linkability of transactions, which makes de-anonymisation inference attacks possible. There have been multiple research works conducted on blockchain transaction analysis, however, these have differing focuses and they usually do not involve the events that are emitted by the transactions in their analysis. Analysis in regards to NFTs is mainly limited to their market analysis and related works do not consider their actual usage in dApps. For instance, Tao et al.~\cite{9614336} presented a network analysis of the Bitcoin blockchain, and Pelechrinis et al.~\cite{10.1371/journal.pone.0287262} provided an analysis of the NBA's TopShot NFT marketplace regarding illegal activities such as money laundering or trading illicit goods. However, the type of user behaviour that can be extracted from dApps through transaction analysis remains unknown which prompted this research as there is a need to present the full extent of the previously mentioned privacy concern. Furthermore, behaviour analysis can be leveraged to identify malicious entities such as phishing scammer participants~\cite{GHOSH2023100153}.

This work aims to analyse behavioural information regarding in-app activities, including NFT usage by users of blockchain-based games. We argue that this type of information can be considered problematic, as malicious users can use it to their advantage, for example, for impersonation attacks or targeted scams~\cite{CHEN2022100048,9169433}. It can also be used by related parties such as game providers, to track the users and their activities and implement changes based on the corresponding analysis. Tracking users in blockchain networks can be done in multiple application areas. Hu et al.~\cite{HU2024110360} studied tracing users for blockchain data audit purposes.

In our work, a blockchain-based game called Planet IX\footnote{https://planetix.com/} was chosen as a case study application. Transactions from multiple smart contracts were collected and used to get the general transaction information and the emitted event logs. The event logs were decoded to obtain all the event data which was then converted into a graph format, thereby allowing it to be queried to access information easily. All related event sequences of users and NFTs were extracted through database queries, which allowed the formation of general in-game actions. From the event sequences and formed actions, the corresponding user action flows for every user were presented. Similarly, for every tokenId (which is a property that corresponds to an NFT), the associated actions were extracted. They were also linked to user wallet addresses, which enabled the presentation of extended user flows that not only show the conducted activities but also which tokens were used at what time by the performed actions.

These extended user flows served as input for a GNN model that assigned unique graph embeddings for each of them. A graph embedding is a vector representation of the graph that is suitable to be used for clustering. Following that, several clustering algorithms were utilised to cluster the user flows into separate unique behavioural clusters. The algorithms were evaluated based on common metrics, and one of them was selected as the current method for the clustering process. The output behavioural clusters were then described which presented the separate types of user behaviours that are possible in this chosen blockchain-based game. A corresponding privacy threat model was also included to present how this can potentially be utilised maliciously.

This research presents a new type of blockchain analysis that focuses on behavioural information that can inform users' activities, interactions and habits. The malicious actors or dApp developers do not need to fully identify the person behind the user as they can construct a profile based on the behavioural information and use that to their advantage to conduct activities like targeting the original user based on past actions or using the constructed profile to perform impersonation-related attacks. Behavioural information does not directly affect users' privacy but by linking multiple activities, relations and habits of users together, it can reveal sensitive information. Therefore, behavioural information is privacy-sensitive and has to be protected. The overall contributions of this research are summarised as follows:

\begin{itemize}
    \item Proposed an action synthesis method that presented unique actions from a blockchain-based application which were used to construct users' action flows.
    \item A user behaviour analysis pipeline was proposed that adopts GNN. Multiple clustering algorithms were utilised within the pipeline to cluster the user flows into unique behavioural clusters. These clusters can be used to analyse different user groups.
    \item The results were compared from the clustering algorithms in order to identify the well performing ones for this task. Descriptions were provided of the different behavioural clusters that a chosen clustering algorithm generated. A privacy threat model was also presented that corresponded to the extracted behaviour clusters and was linked to potential application areas.
\end{itemize}

The rest of the paper is constructed as follows: Section~\ref{sec:related_works} presents related research works. In section~\ref{sec:pipeline}, the proposed user behaviour analysis pipeline is introduced in full detail. Section~\ref{sec:results} shows the clustering results from the chosen clustering algorithm and describes the privacy threat model. Finally, the paper is concluded in section~\ref{sec:conclusion}.
\section{Related work}
\label{sec:related_works}
In this section, the corresponding related works on blockchain transaction analysis, blockchain-based games and the usage of GNN are presented to highlight the differences and limitations in this area of research.

Transaction analysis has been utilised for multiple purposes in previous research. Wu et al.~\cite{wu2022tracer} proposed a novel transaction tracing tool for account-based blockchains called TRacer to trace illicit financial flows. They formed the accounts and their token transfer relationships into a graph format, which enables them to model Decentralised Finance (DeFi) actions. Those were categorised into two patterns: Xfer (transfer, minting, burning) and Swap (add liquidity, remove liquidity, trade), where Xfer refers to sending/receiving tokens and Swap means the exchange of a token for another token. Bonifazi et al.~\cite{bonifazi2022} proposed a new approach to classify Ethereum users. They built a social network that included the user addresses and their transactions, introduced multiple features to characterise the addresses, and utilised multivariate time series. Four classes of interest from information provided by Etherscan were also defined: Token Contract class involves addresses using tokens and not Ether. The Exchange class includes the addresses that buy and sell cryptocurrencies. The Bancor class is where users who allow clients to deposit and convert belong. Finally, the Uniswap class includes users who use the Uniswap protocol.

There has been research that focused on blockchain-based games. Jiang et al.~\cite{10.1145/3494106.3528675} investigated how blockchain can introduce advantages regarding loot boxes. These boxes are goods involving a probability of obtaining one/more in-game virtual assets. They modelled in-game interactions between the provider and the players as a two-stage Stackelberg game. However, they utilised that to conduct market analysis, for example, the price set by the game provider, the users' utility and the impact of the gas fee. Jiang et al~\cite{10.1145/3555858.3555883} collected a year of player operation data from a blockchain-based game called Aavegotchi. They analysed this data in aspects of gaming and finance. The analysis includes daily active addresses, functions and density distribution to reflect player behaviours. Finally, they applied an unsupervised Self-Organising Map (SOM) algorithm to divide user groups into different clusters. Their work presented a similar aim as the behavioural clusters in this work, as they provided information on user activity levels and included the functions from the smart contracts in their data collection. However, when presenting their clusters, they mainly focused on describing how much the corresponding users would invest. They included their interactions but not in detail and did not provide graph visualisation.

GNN is leveraged in the proposed pipeline. It can be applied to multiple research tasks as it is shown as follows. Wu et al. presented a survey on how GNN can be applied to recommender systems \cite{10.1145/3535101}. They presented multiple areas where they can be utilised such as in user-item collaborative filtering. They apply it to exploit high-order connectivity from user-item interactions, or in social recommender systems. They use GNN to model social interactions. Jin et al.~\cite{jin2023survey} similarly showed how GNN is being used in time series-related tasks, like classification and anomaly detection, as they are able to capture inter-variable and inter-temporal relationships. They can be applied to blockchain-related use cases as well. For example, Liu et al.~\cite{9756314} proposed a novel GNN framework, Filter and Augment Graph Neural Network, to provide an Ethereum transaction network embedding model. In the proposed pipeline, user flows are extracted, and they are subgraphs of the second local database's entire graph data. Alsentzer et al.~\cite{NEURIPS2020_5bca8566} introduced SUBGNN that can learn subgraph representations, including if they are disentangled. They also detailed the challenges of utilising GNNs in regard to subgraphs, which we encountered when applying a GNN model to provide subgraph embeddings.

\paragraph{The gap:}
As presented, other blockchain analysis works may include the introduction and usage of actions in analysis, however, the application area is very different. Mainly finance-related. They can also introduce classes that are also a categorisation of the users, similar to the behavioural clusters that are introduced in this work, however, they do not have a direct relation to in-game activities. The proposed pipeline can be beneficial in describing user behaviour in multiple application areas as it can use data from various types of blockchain-based applications. For example, these clusters are based on in-game data, and as a result, they are less general and can be more expressive in the area of gaming.

Works on blockchain-based games are often market-related analyses and cannot be applied directly to this work's research area as they do not present user behaviour. Even if they include user categorisation in classes or clusters, the main focus still relies on potential investment in the game or other dApp and not on general application activities. This research addresses this gap by introducing a novel pipeline that can be used to describe different user behaviours, which can focus on non-finance-related areas as well such as NFT usage or which parts of the game are popular.
\section{User behaviour analysis pipeline}
\label{sec:pipeline}
\begin{figure*}[htbp]
\centerline{\includegraphics[width=\textwidth]{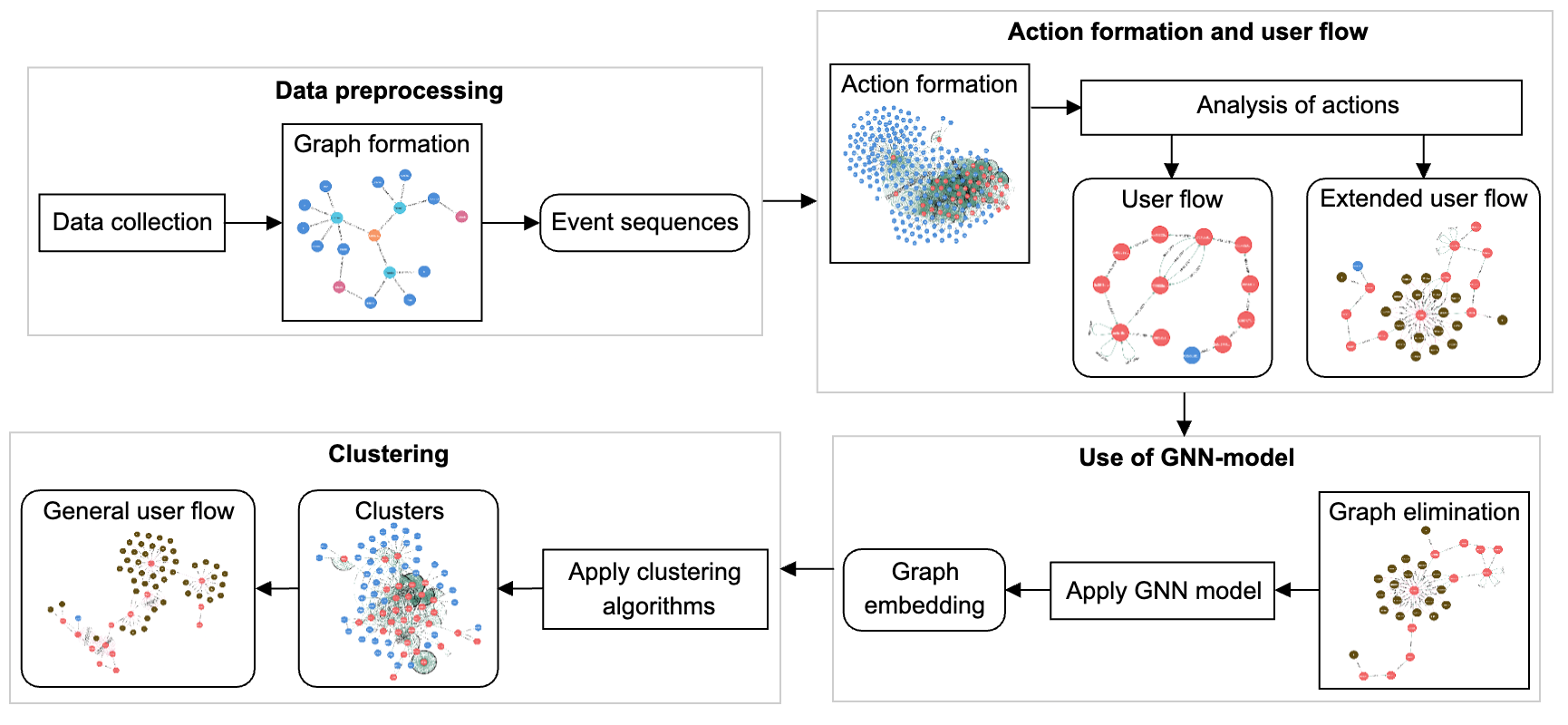}}
\caption{An overview of user behaviour analysis pipeline.}
\label{fig:pipeline}
\end{figure*}
In this section, the introduced user behaviour analysis pipeline is presented that can be used to reveal the type of behavioural information that can be extracted through transaction analysis. As can be seen in Figure~\ref{fig:pipeline}, it has four components: data preprocessing, action formation and application, GNN and clustering. First,
an overview of the pipeline is given and then each component is detailed.

\subsection{Overview}
Each step of Figure~\ref{fig:pipeline}, namely Data preprocessing, Action formation and application, GNN model, and Clustering, is briefly described in this section to give a general overview. These four are explained in detail later in their corresponding sections. The analysed outcomes are presented in section~\ref{sec:results}.

\paragraph{Data preprocessing:}
Transaction information is collected and stored in a graph format in a local database. By querying the graph database, event sequences can be extracted for user wallet addresses. There are two ways of extracting events that correspond to a certain wallet address: enquiring events from transactions that were submitted by the particular user who has that wallet address or extracting events based on event property information that corresponds to the address.

\paragraph{Action formation and user flow:}
The event sequences were examined, and certain patterns were identified that were used to form unique in-game actions. For every sequence, the pattern they belong to was checked, and the corresponding action steps were also added and associated with the particular user's wallet address. An action step refers to one in-game step a user makes at a given time that is represented by a timestamp where the user performs a specific action. The newly formed actions and the noted action steps enabled us to establish a new local database with two newly introduced node types: users and actions. The action steps are presented as edges between these nodes. By converting the data into a graph format, it is possible to present how and in which order the users performed these unique actions in the form of user flows. A similar process can be used to extract event sequences for NFT tokenIds. The tokenIds are unique numbers that identify an NFT within a collection. Querying the first local database, by using the particular tokenId as an event property value, lead to the extraction of the event sequences. After that, all the sequences were checked and assigned to the existing action patterns or formed new actions. The token action steps were also added. Every added action step included the wallet address of the user that was associated with the particular step (for example, the user that submitted the corresponding transaction). This enabled us to add the NFTs as nodes and the token action steps as a new type of edge in the second database. With this new information, not only the user's action activities but also their NFT usage in the form of an extended user flow can be presented. A second local database was added where the action-related information was stored as graphs.

\paragraph{GNN model:}
By examining multiple user flows, it can be seen that some users conduct very similar activities, which indicates that there is a possibility of analysing similar user behaviours if they are grouped together. In this research, this is formulated as an unsupervised clustering task. As a first step, a GNN model was leveraged to provide graph embeddings for its input graph representation. The extended user flows were passed as inputs for this model, however, there was an elimination step to reduce complexity. In the GNN model, both the node and edge features were considered. From the included wallet addresses 70\% of them were randomly chosen for the training, and their corresponding user flows were added as input data. The remaining addresses were used for testing.

\paragraph{Clustering process:}
The elbow method \cite{thorndike1953belongs} was used to estimate the number of clusters that were required because we did not have any predefined knowledge or labels. From the previous GNN step, we had two embedding arrays that were used as input in the clustering process. Multiple clustering algorithms were leveraged, and then their results were compared through multiple clustering metrics. One of the algorithms was chosen based on the comparison, and the different behaviours coming from the corresponding clusters were described and presented. Following that, the privacy threat model was also presented.

In what follows the technical details of the above approach are presented.

\subsection{Data preprocessing}
\label{sec:data_preprocessing}
The first step of the methodology is described which includes the process of the data collection and how that transaction information is converted into graph format which then enables its analysis through various queries. The second part of this describes how the event sequences for the users and the NFTs can be extracted.

\begin{figure*}[h!]
    \centering
    \begin{minipage}{0.4\textwidth}
        \begin{subfigure}[b]{\textwidth}
            \centering
            \includegraphics[width=\textwidth]{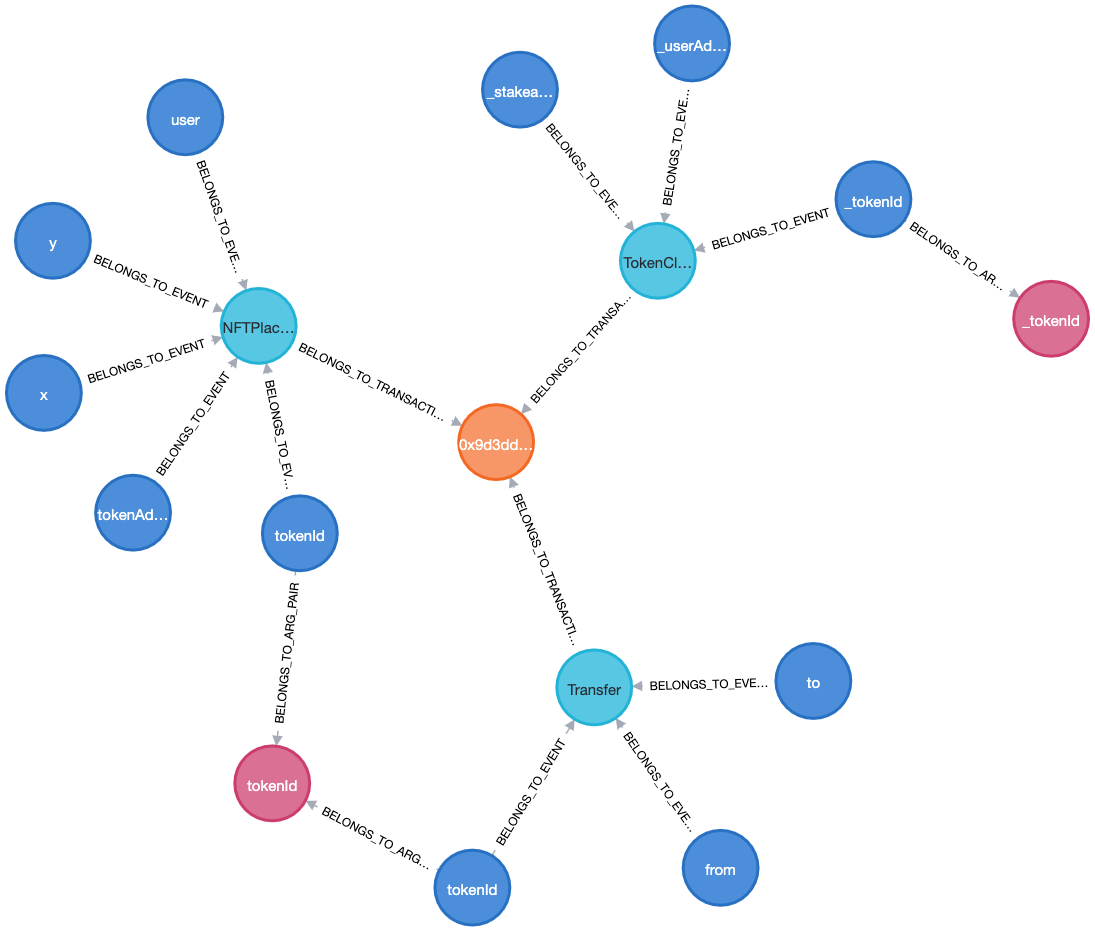}
            \caption{One-time NFT usage}
            \hspace{10pt}
            \label{fig:one-time}
        \end{subfigure}
    \end{minipage}
    \hfill
    \begin{minipage}{0.4\textwidth}
        \begin{subfigure}[b]{\textwidth}
            \centering
            \includegraphics[width=\textwidth]{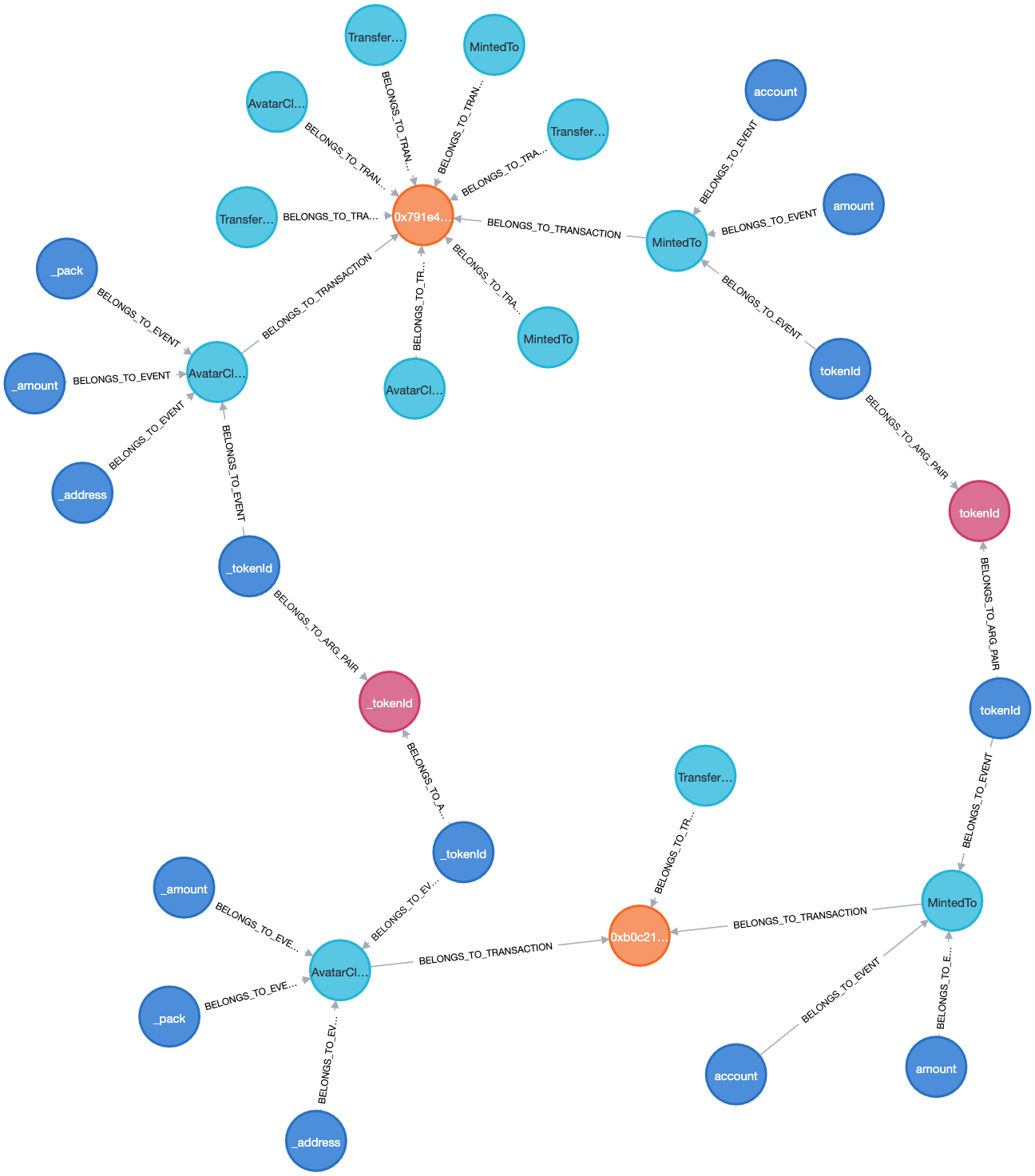}
            \caption{Multi-time NFT usage}
            \label{fig:multiple-time}
        \end{subfigure}
    \end{minipage}
    \hfill
    \begin{minipage}{0.17\textwidth}
        \begin{tikzpicture}
            \filldraw[fill={rgb,255:red,255; green,165; blue,0}, draw=none] (0, 0) circle (0.1cm);
            \node[right] at (0.3, 0) {Transactions};
            \filldraw[fill={rgb,255:red,135; green,206; blue,235}, draw=none] (0, -0.6) circle (0.1cm);
            \node[right] at (0.3, -0.6) {Evens};
            \filldraw[fill={rgb,255:red,0; green,125; blue,255}, draw=none] (0, -1.2) circle (0.1cm);
            \node[right] at (0.3, -1.2) {Event properties};
            \filldraw[fill={rgb,255:red,219; green,112; blue,147}, draw=none] (0, -2) circle (0.1cm);
            \node[right, align=left] at (0.3, -2) {Unique event \\ property pairs};
        \end{tikzpicture}
    \end{minipage}
    \caption{Categorisation of NFTs based on usage (unique event property pairs link event property information across multiple events).}
    \label{fig:nft-usages}
\end{figure*}

\subsubsection{Data collection and graph formation}
For this study, transaction data from a blockchain-based game called Planet IX was collected. Based on our previous work \cite{10.1007/978-981-97-0006-6_6}, APIs were leveraged, namely Polygonscan\footnote{https://docs.polygonscan.com/} and Alchemy\footnote{https://docs.alchemy.com/reference/api-overview}, to extract data from the Polygon blockchain where the game is deployed. In addition to the basic transaction information (such as transactionHash, blockHash and value), the events that are emitted within the transactions are also encoded. The collected data is then transformed into graph format and stored in a local Neo4j database. This enables the process of conducting blockchain transaction analysis on the collected game data, which can reveal connections between transactions, events and event properties. 

The data coming from the emitted events includes various types of information such as token and staking-related properties. It also presents a large number of involved wallet addresses which are the potential user addresses that can be examined to reveal user behaviour. By querying the established local database 12146 distinct addresses were extracted. However, this number is a combination of user, smart contract and game-related addresses.

\subsubsection{Event sequencing}
Since in-game behavioural information~\cite{10.1145/3402942.3402962} is extracted, event sequences have to be linked to addresses. When building these sequences, the following has to be taken into account: the events that were emitted through the transactions that were submitted by the particular address, and the events that were part of other transactions, however, the specific wallet address is included as event property information. The latter transactions could have been initiated by a user as the differing wallet address can be just the result of the structure of the game, therefore, it has to be considered. For the first type of event sequence, the database can be queried based on the from address transaction information which corresponds to the address that submitted the transaction. The second type of event sequence is extracted by querying connections between events through event property information that has the address as the value. Besides the addresses' event sequences, the sequences that correspond to NFTs are also extracted, however, in this case, only the second type of event sequences are possible because the query is based on the tokenIds of the NFTs. Following that, both the user and NFT event sequences are ordered based on the timestamp transaction information. By analysing these sequences, the type of activities the users or NFTs have been involved with can be learned.

When looking into the NFT event sequences, it can be seen that certain NFTs are only leveraged once by one address, whereas others can be part of several events and can be leveraged by multiple addresses. An example of the one-time NFT usage is given in Figure~\ref{fig:one-time}, where an NFT is placed down to a certain location and is not referenced again. Figure~\ref{fig:multiple-time} presents multiple-time NFT usage for a tokenId that represents an avatar that is claimed multiple times. We have to note, however, that a tokenId does not fully describe an NFT, as multiple collections can have the same tokenId value, but this still can indicate multiple usage of the same NFT. In the future, this is planned to be extended. 

\subsection{Action formation and user flow}
\label{sec:actions}
The formation of the unique in-game actions from the address event sequences is presented. Two types of actions are introduced: primary and secondary. It also describes how the corresponding data can be converted into graph format. Following that, the way to present the corresponding user flows derived from the event sequences and also the method to extend them with the inclusion of the NFTs will be explained. 

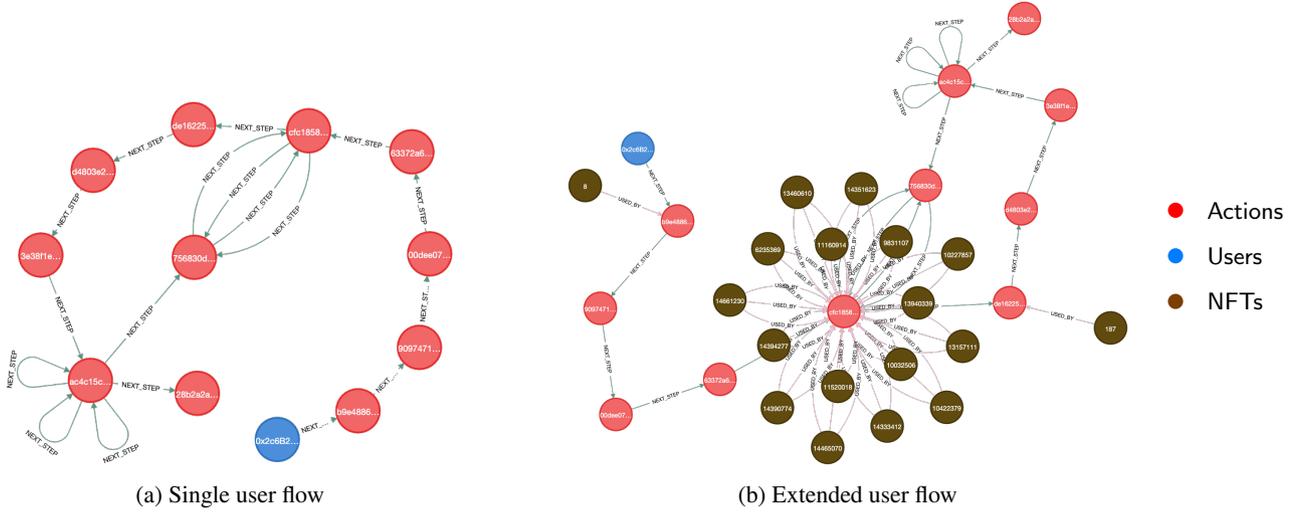
\begin{figure*}[h!]
    \centering
    \begin{minipage}{0.85\textwidth}
        \begin{subfigure}[b]{0.4\textwidth}
            \centering
            \includegraphics[width=\textwidth]{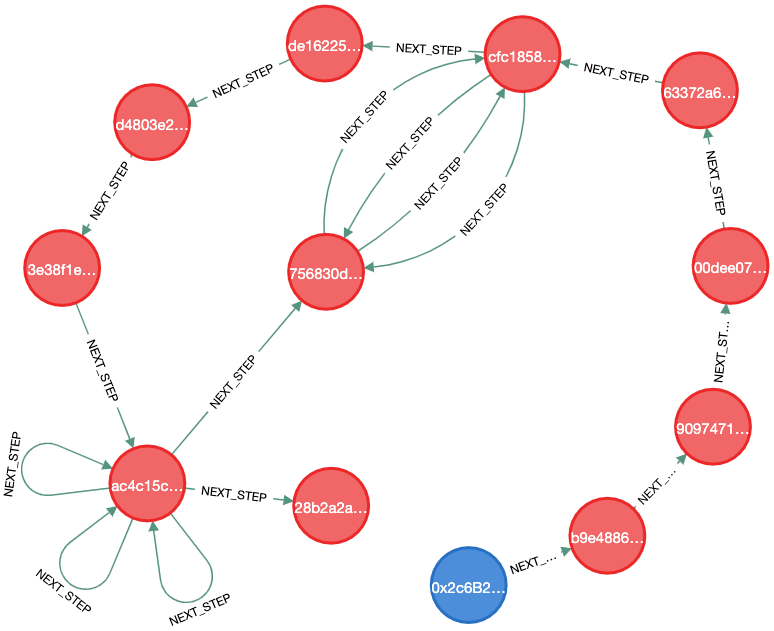}
            \caption{Single user flow}
            \label{fig:user-flow-example}
        \end{subfigure}
        \hfill
        \begin{subfigure}[b]{0.5\textwidth}
            \centering
            \includegraphics[width=\textwidth]{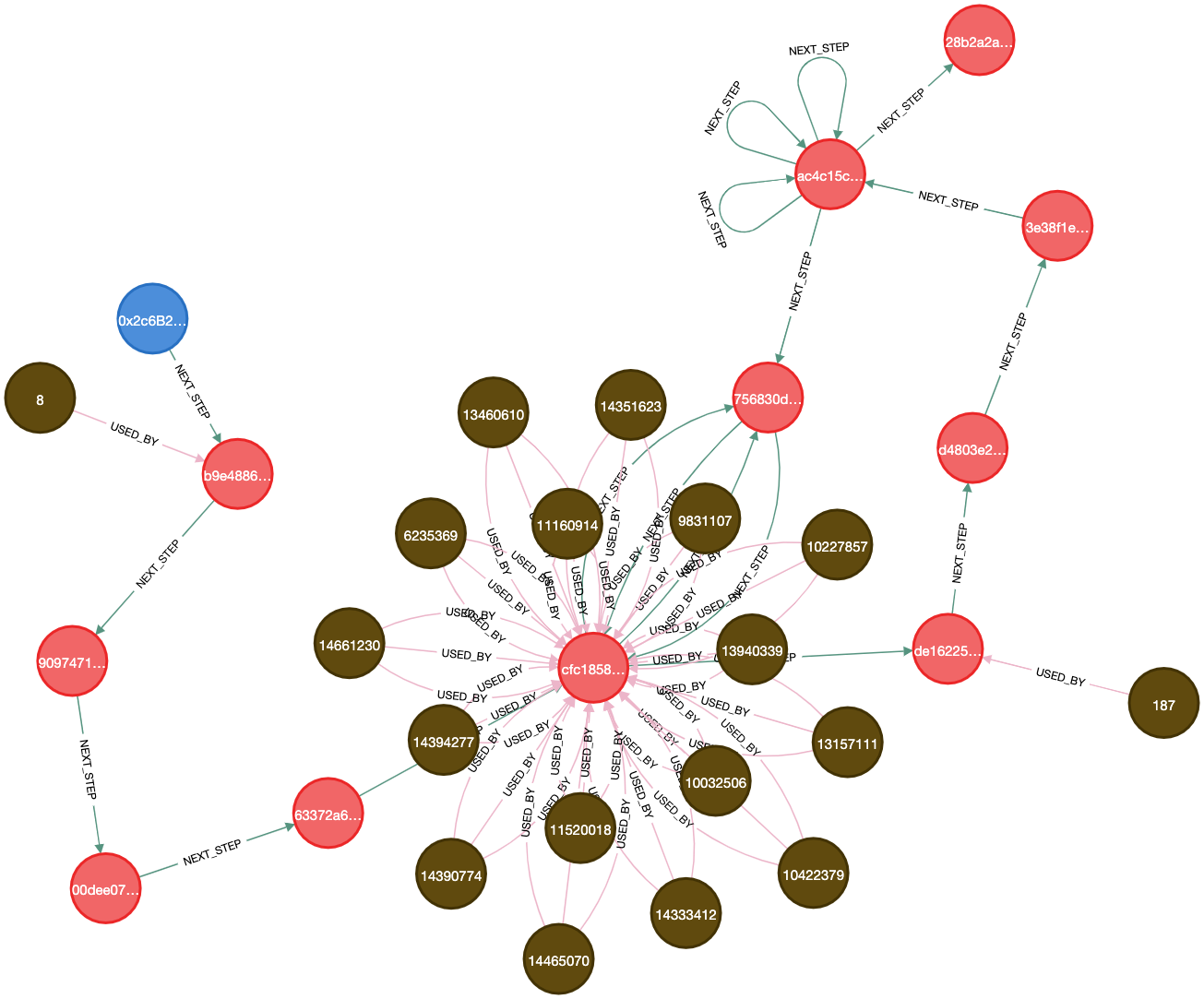}
            \caption{Extended user flow}
            \label{fig:extended-user-flow-example}
        \end{subfigure}
    \end{minipage}
    \hfill
    \begin{minipage}{0.12\textwidth}
        \begin{tikzpicture}
            \filldraw[fill={rgb,255:red,255; green,0; blue,0}, draw=none] (0, 0) circle (0.1cm);
            \node[right] at (0.3, 0) {Actions};
            \filldraw[fill={rgb,255:red,0; green,125; blue,255}, draw=none] (0, -0.6) circle (0.1cm);
            \node[right] at (0.3, -0.6) {Users};
            \filldraw[fill={rgb,255:red,123; green,63; blue,0}, draw=none] (0, -1.2) circle (0.1cm);
            \node[right] at (0.3, -1.2) {NFTs};
        \end{tikzpicture}
    \end{minipage}
    \caption{Examples of user flows.}
    \label{fig:user-flows}
\end{figure*}

\begin{algorithm}
\caption{Action formation from event sequences}
\label{alg:alg_actions}
\begin{algorithmic}[1]
    \Procedure{Form actions}{$type, addresses, patterns$}
        \State $sequences \gets []$
        \State $actions \gets []$
        \State $action\_steps \gets []$
        \For{$\textit{addr}$ in $\textit{addresses}$}
            \If{$\textit{type} = \text{primary}$}
                \State $\textit{sequences} \gets$ primary event sequences
            \Else
                \State $\textit{sequences} \gets$ secondary event sequences
            \EndIf
            \State $order \gets 0$
            \For{$\textit{seq}$ in $\textit{sequences}$}
                \State $\textit{current\_pattern} \gets$ \textbf{call} 
                match\_pattern ($seq, patterns$)
                \State $\textit{current\_events} \gets$ \textbf{call} 
                get\_events ($seq.events, current\_pattern$)

                \For{each \textit{action} in \textit{actions}}
                    \If{\textit{action.events} $==$ \textit{current\_events}}
                        \State $\textit{uuid} \gets \textit{action.id}$
                    \Else
                        \State $\textit{uuid} \gets$ generate uuid
                        \State \textit{new\_action} $\gets$ \{\textit{uuid}: \textit{uuid}, \textit{type}: \textit{type}, \textit{events}: \textit{current\_events}\}
                        \State \textit{actions.push(new\_action)}
                    \EndIf
                    \State \textit{new\_step} $\gets$ \{\textit{order}: \textit{order}, \textit{uuid}: \textit{uuid}, \textit{prev\_uuid}: \textit{\textit{action\_steps}[\textit{length}(\textit{action\_steps}) - 1]}.uuid, \textit{address}: \textit{addr}, \textit{transactionHash}: \textit{seq.transactionHash}, \textit{timestamp}: \textit{seq.timestamp}, \textit{data}: \textit{seq.data}\}
                    \State \textit{action\_steps.push(new\_step)}
                \EndFor
                \State \textit{order} $\gets$ \textit{order} + 1
            \EndFor
        \EndFor
        \State \textbf{return} $actions, action\_steps$
    \EndProcedure
\end{algorithmic}
\end{algorithm}

\subsubsection{Formation of actions}
In section~\ref{sec:data_preprocessing}, two types of event sequences were identified. The first type extracts them from the transactions that were submitted by the wallet address. Actions that are derived from these sequences were named as primary actions since, in this case, it is certain that they describe address-related activities. The second type acquires the events based on connecting events through the address as a common event property. Every action that is constructed from these sequences, is secondary. Even though the address is included as event information, the action may or may not describe an activity the actor behind the address personally has been a part of. Some of the identified actions can be considered both primary and secondary as they were derived from both types of event sequences. We identified 48 actions where 38 of them were primary, one was secondary and 9 were determined as both.

When examining the transactions the events were extracted from, regardless of which case the transactions were considered in, it can be seen that the events are repeating across multiple similar transactions in unique recognisable patterns. These patterns were identified and used to extract the unique actions that all actors conduct within the game. These patterns are explained as follows:

\begin{enumerate}[leftmargin=3ex]
  \item Unique events: This corresponds to those transactions that have a unique set of events emitted that is fixed at every occurrence. This transaction is submitted by a wallet address.
  \item Every event repeating the same amount of times: This presents transactions where, within the event set, event sequence chunks can be discovered that consist of the same set of events. The event sequence chunk is added as the new action. The sequence chunks can be clearly grouped based on their event property values.
  \item Every event repeats the same amount of times except for one event, which only occurs once: This is similar to the previous pattern except for
  the single count event. It is recognised that the latter event is actually a bulk event, which means that it includes event property values from the event sequence chunks initiated from the repeating events. This is added as a new action formed as the combination of the event sequence chunk and the bulk event.
  \item One event repeats a large number of times, whereas the rest of them only occur once: In the corresponding transactions, it can be noticed that there is one event that occurs a large but varying number of times, while other events are unique per transaction. This is added as a new action by only adding the repeating event once but by combining the event property values.
  \item Some events repeat at the same amount of times, but there are single count events as well: This groups all the remaining other transactions as there are no recognisable patterns. In this case, the actions are added with all the events included.
\end{enumerate}

These unique actions were added by going through the event sequences. For every sequence, it was first checked which pattern they belonged to. Following that, the then-current action list (as the actions were formed, the action list was continuously growing) was examined to investigate whether the current sequence corresponded to an already identified action. If not, the sequence was added as a new action with a generated uuid (unique ID that is generated and then assigned to the action), the action type (primary, secondary or both) and the associated events. The action step was also noted with the following information: The order of the step (the order of the steps were able to be ascertained because the sequences were previously ordered based on the timestamps), the corresponding wallet address (in the NFTs' case this is the from address), in case of an NFT-related sequence the tokenId, the transactionHash, the timestamp, the previous and current action uuids and the event data information. In the latter, the tokenIds or IDs for the assets (NFTs), tickets (the game includes a lottery activity) and packs (openable objects that include additional items like avatar NFTs), the related addresses, the values (such as reward amount) and additional IDs (like IDs associated with waste management which is one of the in-game activities) were noted. In the NFT's case, only the related addresses were noted. The action formation from sequences can be seen in Algorithm~\ref{alg:alg_actions} where the function \textit{match\_pattern} matches the events with the established patterns and function \textit{get\_events} shortens the event list based on the matched pattern.

\begin{table}[]
\caption{Properties of the action nodes.}
\label{tab:action-properties}
\resizebox{0.48\textwidth}{!}{%
\begin{tabular}{|l|l|}
\hline
\multicolumn{1}{|c|}{\textbf{Property}}      & \multicolumn{1}{|c|}{\textbf{Description}}                                                        \\ \hline
Total count   & total \# times action is called               \\ \hline
Min call      & min \# times action is called               \\ \hline
Max call      & max \# times action is called               \\ \hline
Mean call     & avg \# times action is called               \\ \hline
Min assets    & min \# assets associated with action            \\ \hline
Max assets    & max \# assets associated with action            \\ \hline
Mean assets   & mean \# assets associated with action            \\ \hline
Min tickets   & min \# tickets associated with action            \\ \hline
Max tickets   & max \# tickets associated with action            \\ \hline
Mean tickets  & mean \# tickets associated with action            \\ \hline
Min packs     & min \# packs associated with action            \\ \hline
Max packs     & max \# packs associated with action            \\ \hline
Mean packs    & mean \# packs associated with action            \\ \hline
Min timestamp & first time action is called                          \\ \hline
Max timestamp & last time action is called                          \\ \hline
\end{tabular}%
}
\end{table}

\subsubsection{Graph representation}
Similarly to our previous work \cite{10.1007/978-981-97-0006-6_6}, the extracted information was converted into a graph format and stored in a separate local Neo4j database. Three types of nodes were introduced: 1. User nodes that correspond to the unique wallet addresses, as those are the only node property they have. 2. Action nodes that present the unique set of actions that were formed from the event sequences based on the patterns from the previous sub-section. They have multiple properties that were extracted from the examination of the address and NFT action steps. They are summarised those in Table~\ref{tab:action-properties}. 3. NFT nodes that correspond to a unique token and have only one other property (\textit{isMultiple}), which describes whether they are one-time or multiple-time usage NFT. This new local database includes two types of relations: NEXT\_STEP, which presents one step from a specific address at a given timestamp, and USED\_BY, which describes one NFT usage at a given timestamp by a certain address (which can be non-user). Both of them have the same three edge properties: the corresponding wallet address, the order in which the address conducted the action or the NFT was used and the timestamp.

\subsubsection{Extracting user flows}
As mentioned in the previous sub-section, action steps from both the users and the NFTs are converted into a graph format in the form of the NEXT\_STEP and the USED\_BY relations. Since these relationships have the address as a property, the database graph data can be filtered based on that. This results in the user flow graphs that correspond to the filter address only. Since the edges also include the action step order number and the associated timestamps, by examining these user flows, the type of activities the participants have conducted can be described, and assumptions can be made in regard to their behaviour. For example, what time they conduct certain actions, in which order they perform a subset of actions or how active they are in general. Shorter user flows indicate lesser engagement in the game. Figure~\ref{fig:user-flow-example} presents an example of a single user-flow that only shows how the actions have followed each other. If the NFT nodes and their USED\_BY relations are also included in the filtering, an extended user flow can be presented, which also considers how the assets were leveraged by the particular corresponding address. An example of this can be seen in Figure~\ref{fig:extended-user-flow-example}. 

\begin{figure*}[ht!]
\centerline{\includegraphics[width=\textwidth]{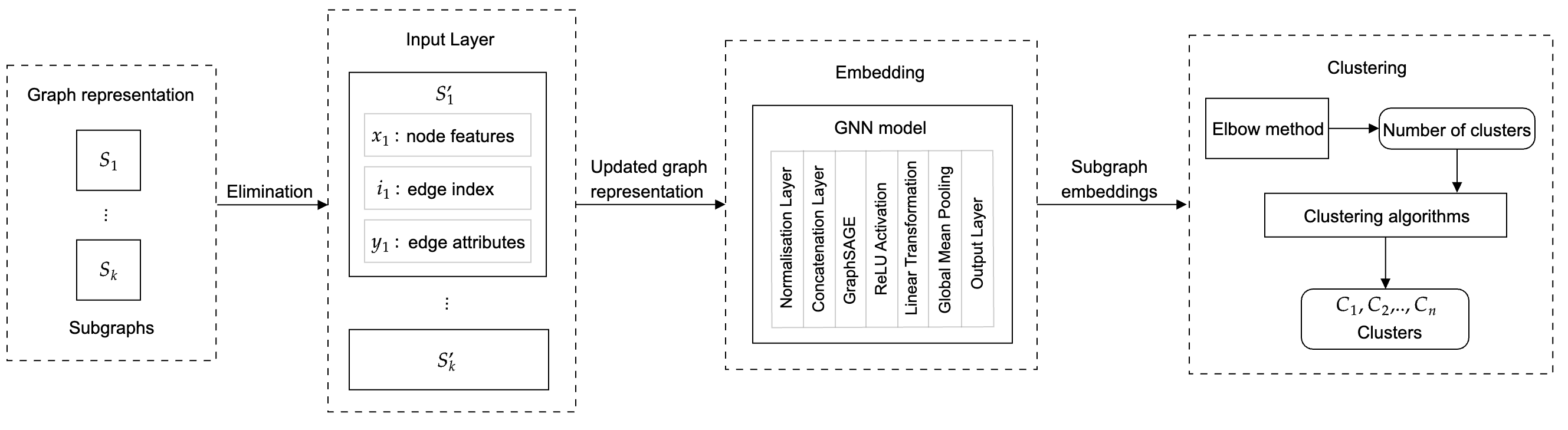}}
\caption{Overall embedding and clustering process.} \label{fig:clustering_process}
\end{figure*}

\subsection{Embedding and clustering}
\label{sec:clusters}
In the following, it is explained how the individual user flows are categorised into multiple separate clusters to show not only a single user behaviour but also to enable the analysis of group user behaviour. For this purpose, a GNN model was included that provides unique embeddings for every user flow. These embeddings are then given as input to the clustering algorithms, which results in clusters for all the user flows and, therefore, enables them to be analysed and visualised together. The entire process is visualised in Figure~\ref{fig:clustering_process}.

\subsubsection{Definition of user flow subgraph and elimination}
Let \( \mathbf{G}(\mathbf{V},\mathbf{E}) \) be a graph with a set $\mathbf{V}$ of vertices and a set $\mathbf{E}$ of edges. There are three types of vertices (nodes): \( \mathbf{A} \subseteq \mathbf{V} \) is the set of action nodes, \( \mathbf{U} \subseteq \mathbf{V} \) is the set of user nodes, and \( \mathbf{N} \subseteq \mathbf{V} \) is the set of NFT nodes. 
For \( \forall \mathbf{u} \in \mathbf{U} \) there is a user flow $\mathbf{S}$,
which is a subgraph of \( \mathbf{G} \).
For every such user flow (sub)graph,
a suitable graph representation \( \mathbf{S'} \) is generated and then given as input for the GNN model to get the corresponding subgraph embedding.

The graph representation \( \mathbf{S'} \) for each $\mathbf{S}$ is being constructed as follows: As 
$\mathbf{S}$ corresponds to one user \( \mathbf{u}\), the related nodes consist of \( \mathbf{V_{\mathbf{S}}} = \{\mathbf{u}\} \cup\{\mathbf{a} \mid \mathbf{a} \in \mathbf{A_u}\} \cup \{\mathbf{n} \mid \mathbf{n} \in \mathbf{N_u}\}\) and \(\mathbf{E} = \{\mathbf{e} \mid \mathbf{e} \in \mathbf{E_u}\}\) 
where the single user node \( \mathbf{u} \) and its associated subset of the action nodes \( \mathbf{A_u} \), subset of the NFT nodes \( \mathbf{N_u} \) and the related edges \( \mathbf{E_u} \) are considered.
The user nodes are emitted from the input graph representations for the GNN model as there is only one user node for every user flow subgraph and the user nodes only have the address property which makes them negligible when it comes to learning. The corresponding edges are also emitted as we assume that by eliminating only one edge, these user flow subgraphs are still suitable for clustering into separate diverse clusters. However, this elimination reduces the complexity of the model because the dimension differences between the user nodes and the other nodes do not have to be handled. Furthermore, every $\mathbf{S}$ that interacted with less than four unique actions is excluded from the embedding and clustering process as it is presumed that an actual user would interact with multiple parts of the game whereas a game-related address will perform the same set of actions continuously. With this step, the included address list is reduced to wallet addresses that are more likely to be actual user addresses. From the extracted 12146 addresses, only 8296 had associated event sequences and solely 716 were considered for clustering. The burn address~\footnote{0x0000000000000000000000000000000000000000} was also eliminated. For the action nodes, the features presented in Table~\ref{tab:action-properties} are considered. The NFT nodes only have the \textit{isMultiple} property to be included. The edge attributes are fully leveraged. Based on this, GNN graph representations were added for the 716 addresses.

\subsubsection{GNN model}
This sub-section presents the GNN model that is leveraged to produce graph embeddings. It consists of multiple layers to provide an embedding that takes into account all the necessary node features and also the edge attributes. By leveraging all, the produced embedding is clearly unique for \( \forall \mathbf{S} \), therefore, is suitable to be given as input for clustering algorithms. 
The following is defined for the model:
\[
\begin{aligned}
& \text{Let } \mathbf{X} \in \mathbb{R}^{N \times d_{\text{in\_node}}} \text{ be the input node features} \\
& \text{Let } \mathbf{Y} \in \mathbb{R}^{M \times d_{\text{in\_edge}}} \text{ be the input edge attributes} \\
& \text{Let } \mathbf{I} \in \mathbb{R}^{2 \times M} \text{ be the edge index} \\
& \text{Let } \mathbf{B} \in \mathbb{R}^{N} \text{ be the batch vector} \\
& \text{Let } \mathbf{H} \in \mathbb{R}^{N \times d_{\text{hidden}}} \text{ be the hidden node embeddings} \\
& \text{Let } \mathbf{Z} \in \mathbb{R}^{G \times d_{\text{out}}} \text{ be the output graph embeddings} \\
\end{aligned}
\]

As users perform different actions in varying order, the sizes of the node features, edge index and edge attributes tensors are different. In the following, all the layers the GNN model consists of are being presented:

\paragraph{Input Layer:} The input consists of node features $\mathbf{X}$, edge index $\mathbf{I}$ and the edge attributes $\mathbf{Y}$. The $\mathbf{X}$ that is given as input is the concatenation of the padded node features matrices of the action and NFT nodes. As they have differing dimensions, this is an addition to reduce the complexity because this way, the GNN model still behaves as a homogenous model instead of the complexity of a heterogenous model. The edge attributes are the same for both relations so there was no need for an adjustment in that regard. The input layer consists of the following:
\[
\mathbf{X} \in \mathbb{R}^{N \times d_{\text{in\_node}}}, \quad \mathbf{I} \in \mathbb{R}^{2 \times M}, \quad \mathbf{Y} \in \mathbb{R}^{M \times d_{\text{in\_edge}}}
\]

\paragraph{Normalization Layer.}
Node features are standardised using layer normalisation \cite{ba2016layer} as a pre-processing step, which was added to improve the performance of the training:
\[
\mathbf{X}_{\text{norm}} = \text{LayerNorm}(\mathbf{X}) \in \mathbb{R}^{N \times d_{\text{in\_node}}}
\]

\paragraph{Concatenation Layer:}
Because the edge attributes are also required to be considered when generating the subgraph embeddings, they are concatenated to the node features. Scatter Mean is applied to calculate the average of the edge attributes and that is then concatenated to the node features. This way all graph information is passed down through the other layers. The applied Scatter Mean is described as follows:
\[
\begin{array}{l}
\mathbf{I}_{\text{edge}} = \text{scatter\_mean}(\mathbf{Y}, \mathbf{I}[0], \text{dim}=0, \\ \text{dim\_size}=N) \in \mathbb{R}^{N \times d_{\text{in\_edge}}}
\end{array}
\]

The concatenated node features $\mathbf{X}_{\text{cat}}$ that is passed down to the GraphSAGE layer is explained as follows:
\[
\mathbf{X}_{\text{cat}} = \text{concat}(\mathbf{X}_{\text{norm}}, \mathbf{I}_{\text{edge}}, \text{dim}=1) \in \mathbb{R}^{N \times (d_{\text{in\_node}} + d_{\text{in\_edge}})}
\]

\paragraph{GraphSAGE:}
 A GraphSAGE~\cite{NIPS2017_5dd9db5e} operator\footnote{https://bitly.cx/oo1b} is utilised to perform the neighbourhood aggregation where every node aggregates features from its neighbouring nodes:
\[
\mathbf{H} = \text{SAGEConv}(\mathbf{X}_{\text{cat}}, \mathbf{I}) \in \mathbb{R}^{N \times d_{\text{hidden}}}
\]

\paragraph{ReLU Activation:}
The previously defined hidden features are passed through a ReLU activation function:
\[
\mathbf{H}_{\text{relu}} = \text{ReLU}(\mathbf{H}) \in \mathbb{R}^{N \times d_{\text{hidden}}}
\]

\paragraph{Linear Transformation:}
A linear transformation is applied to the activated features to put them into the required shape for the next step:
\[
\mathbf{H}_{\text{linear}} = \mathbf{H}_{\text{relu}} \mathbf{W}_{\text{linear}} + \mathbf{b}_{\text{linear}} \in \mathbb{R}^{N \times d_{\text{out}}}
\]

\paragraph{Global Mean Pooling:}
Since the subgraph embeddings from both the training and the testing of this model are required to obtain, global mean pooling was applied to obtain graph-level embeddings:
\[
\mathbf{Z} = \text{global\_mean\_pool}(\mathbf{H}_{\text{linear}}, \mathbf{B}) \in \mathbb{R}^{G \times d_{\text{out}}}
\]

\paragraph{Output Layer:}
Finally, the model returns the total graph-level embeddings for the input $\mathbf{S'}$:
\[
\mathbf{Z} \in \mathbb{R}^{G \times d_{\text{out}}}
\]

\subsubsection{Embedding and clustering of user-flow graphs}
The user (sub)graphs were partitioned into a training and testing dataset. 501 user addresses were chosen, and their corresponding graph representations were constructed and then added to the training dataset. The graph representations were calculated for the remaining 215 users as well and then placed in the testing dataset. The training dataset was used to train the GNN model, and that was validated on the testing dataset by graph embedding similarity. Each unique subgraph embedding was extracted and grouped into one array, which was given as input for the clustering algorithms.

\begin{figure}[ht!]
\centerline{\includegraphics[width=0.55\textwidth]{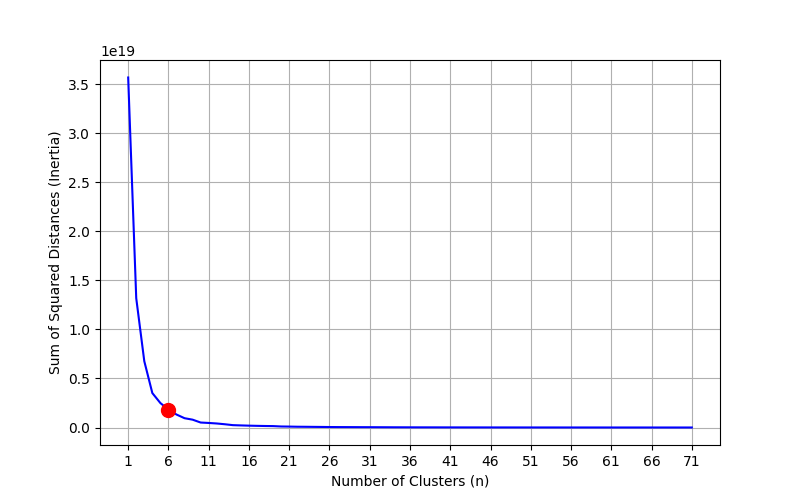}}
\caption{Elbow method for optimal n.}
\label{fig:elbow}
\end{figure}

Without pre-defined knowledge, we also did not have any indicators of the suitable clustering algorithm that should be applied. Therefore, clustering was performed by leveraging multiple clustering algorithms, ranging from widely used to state-of-the-art, and comparing their results. The applied algorithms are as follows: \textit{k}-means \cite{macqueen1967some}, mean-shift \cite{1055330}, spectral clustering \cite{NASCIMENTO2011221}, agglomerative clustering \cite{1427769}, BIRCH \cite{10.1145/233269.233324}, bisecting \textit{k}-means \cite{doi:10.1137/1.9781611972719.5} and affinity propagation \cite{doi:10.1126/science.1136800}. We have also tried DBSCAN \cite{10.5555/3001460.3001507} and HDBSCAN \cite{10.1007/978-3-642-37456-2_14}, but they did not yield comparable results, so we excluded them from our experiment tables.

Prior to this clustering process, we did not have any predefined knowledge of potential user behaviour or any grand truth labels that we could use. Therefore, unsupervised clustering was conducted to place the unique subgraph embeddings into multiple diverse clusters. We also did not have any indication of the potentially required number of clusters (n), so the elbow method was leveraged. Based on Figure~\ref{fig:elbow}, 6 was chosen as the number of clusters for those clustering algorithms that require such an input. 

\begin{table}[]
\caption{Comparison of clustering algorithms.}
\label{tab:clustering-metrics}
\resizebox{0.48\textwidth}{!}{%
\begin{tabular}{|l|c|c|c|}
\hline
\multicolumn{1}{|c|}{\textbf{Algorithm}} & \textbf{SC}     & \textbf{DBI}    & \textbf{CHI}     \\ \hline
\textit{k}-means                                  & \textbf{0.6122} & \hspace{13px}0.4605          & \textbf{2757.17} \\ \hline
Mean-shift                               & 0.5592          & \hspace{13px}0.5456          & \hspace{5px}987.51           \\ \hline
Spectral clustering                      & 0.1463          & \hspace{13px}2.8559          & \hspace{5px}605.45           \\ \hline
Agglomerative clustering                 & 0.5273          & \hspace{13px}0.5712          & 2030.50          \\ \hline
BIRCH                                    & 0.5447          & \hspace{13px}\textbf{0.4321} & 2232.54          \\ \hline
Bisecting \textit{k}-means                        & 0.5685          & \hspace{13px}0.4670          & 2412.91          \\ \hline
Affinity propagation                     & -0.6096\hspace{3px}         & \hspace{5px}646.2999        & \hspace{13px}1.65             \\ \hline
\end{tabular}
}
\end{table}

\begin{figure*}[htbp]
    \centering
    \begin{subfigure}[b]{0.3\textwidth}
        \centering
        \includegraphics[width=\textwidth]{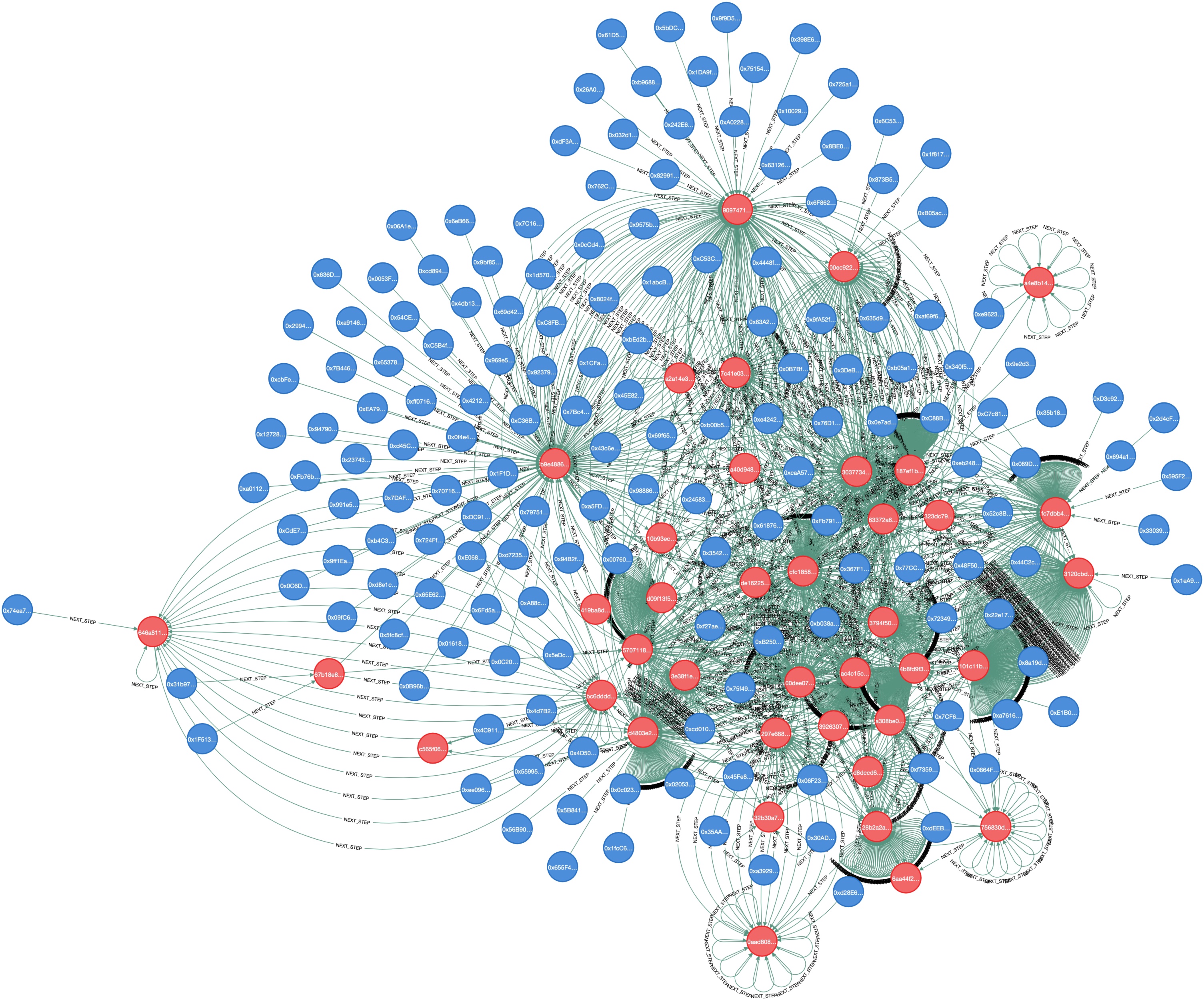}
        \caption{Cluster $C_1$}
    \end{subfigure}
    \begin{subfigure}[b]{0.3\textwidth}
        \centering
        \includegraphics[width=\textwidth]{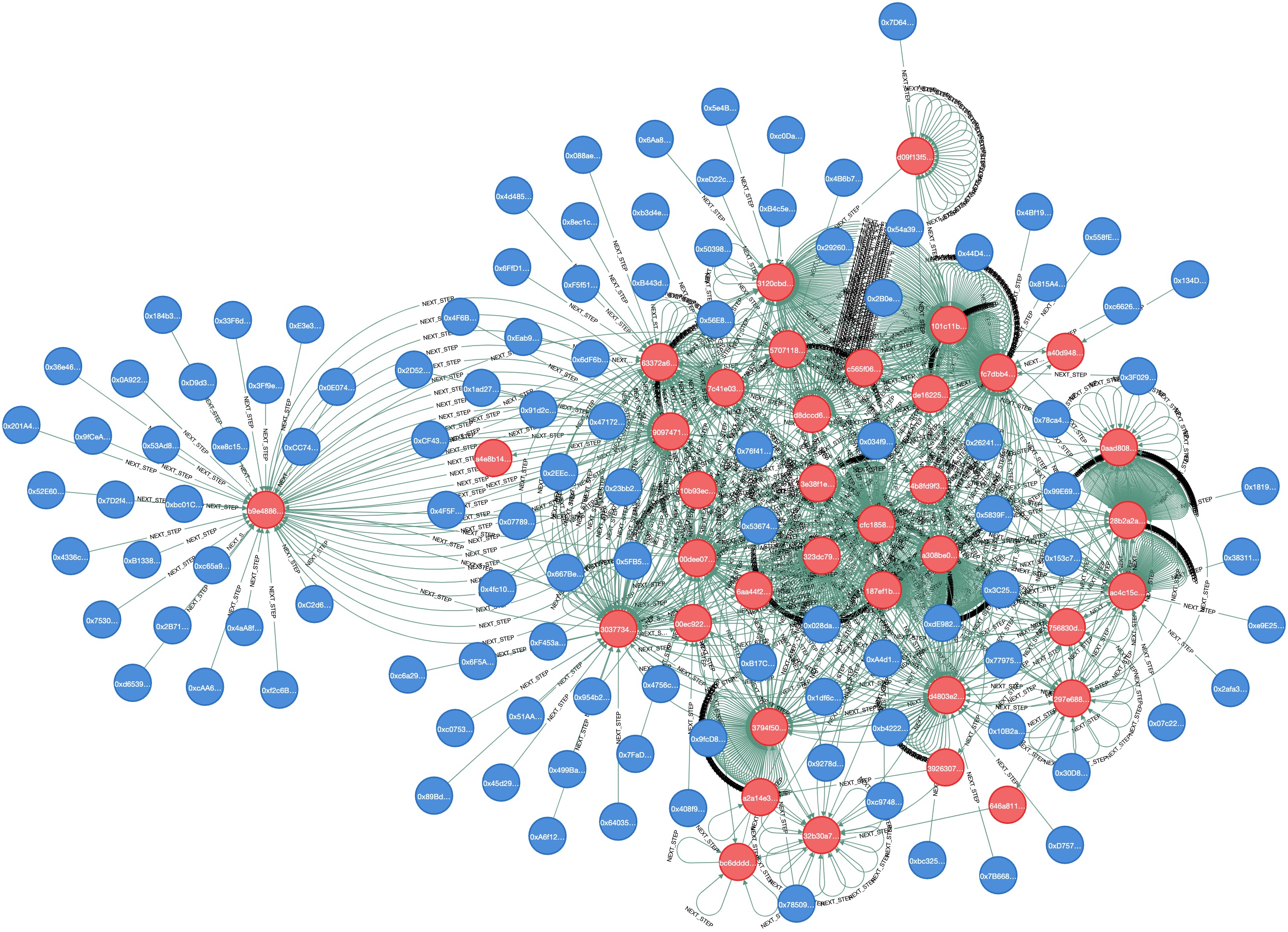}
        \caption{Cluster $C_2$}
    \end{subfigure}
    \begin{subfigure}[b]{0.3\textwidth}
        \centering
        \includegraphics[width=\textwidth]{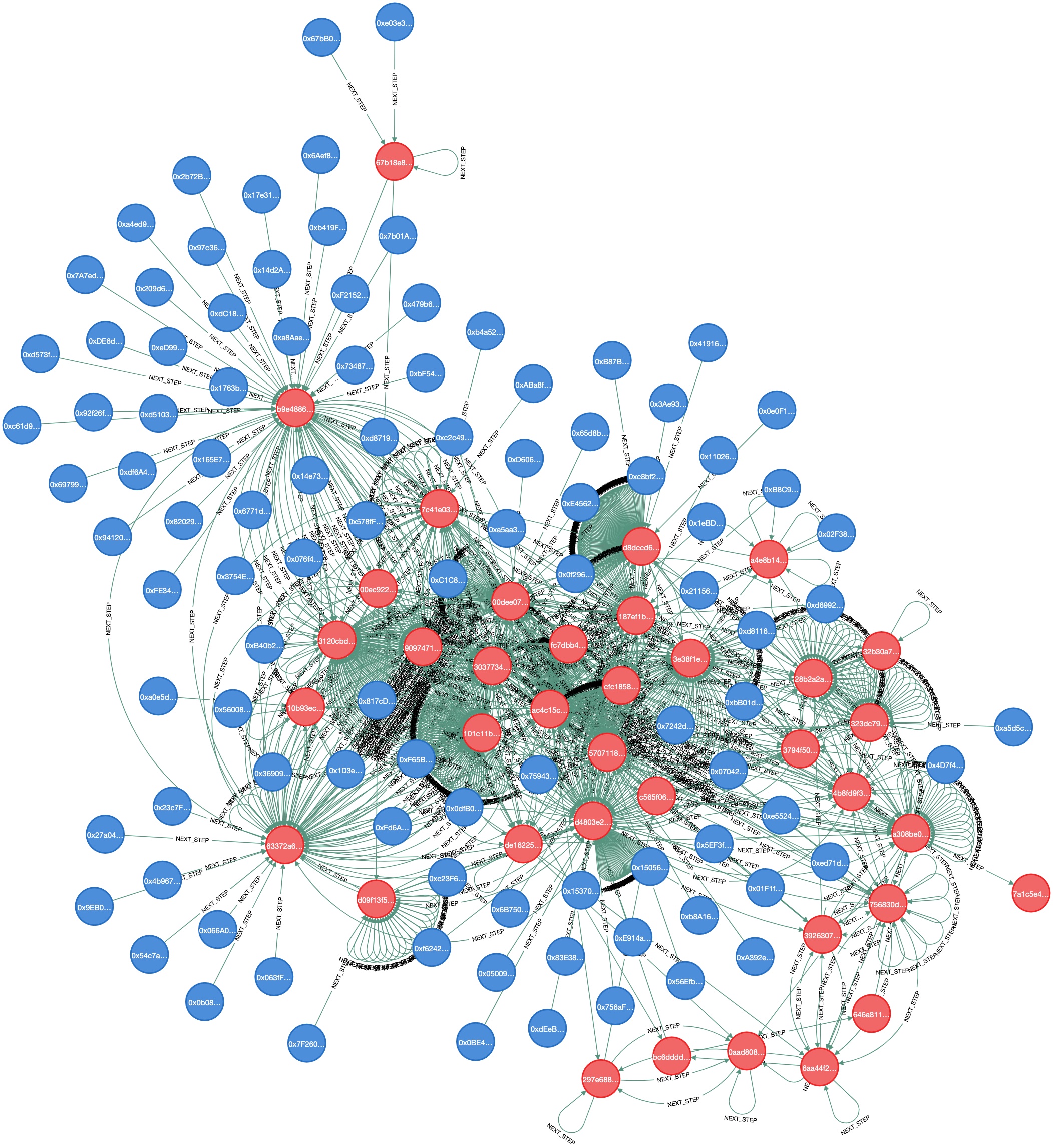}
        \caption{Cluster $C_3$}
    \end{subfigure}
    
    \vspace{1cm}
    
    \begin{subfigure}[b]{0.3\textwidth}
        \centering
        \includegraphics[width=\textwidth]{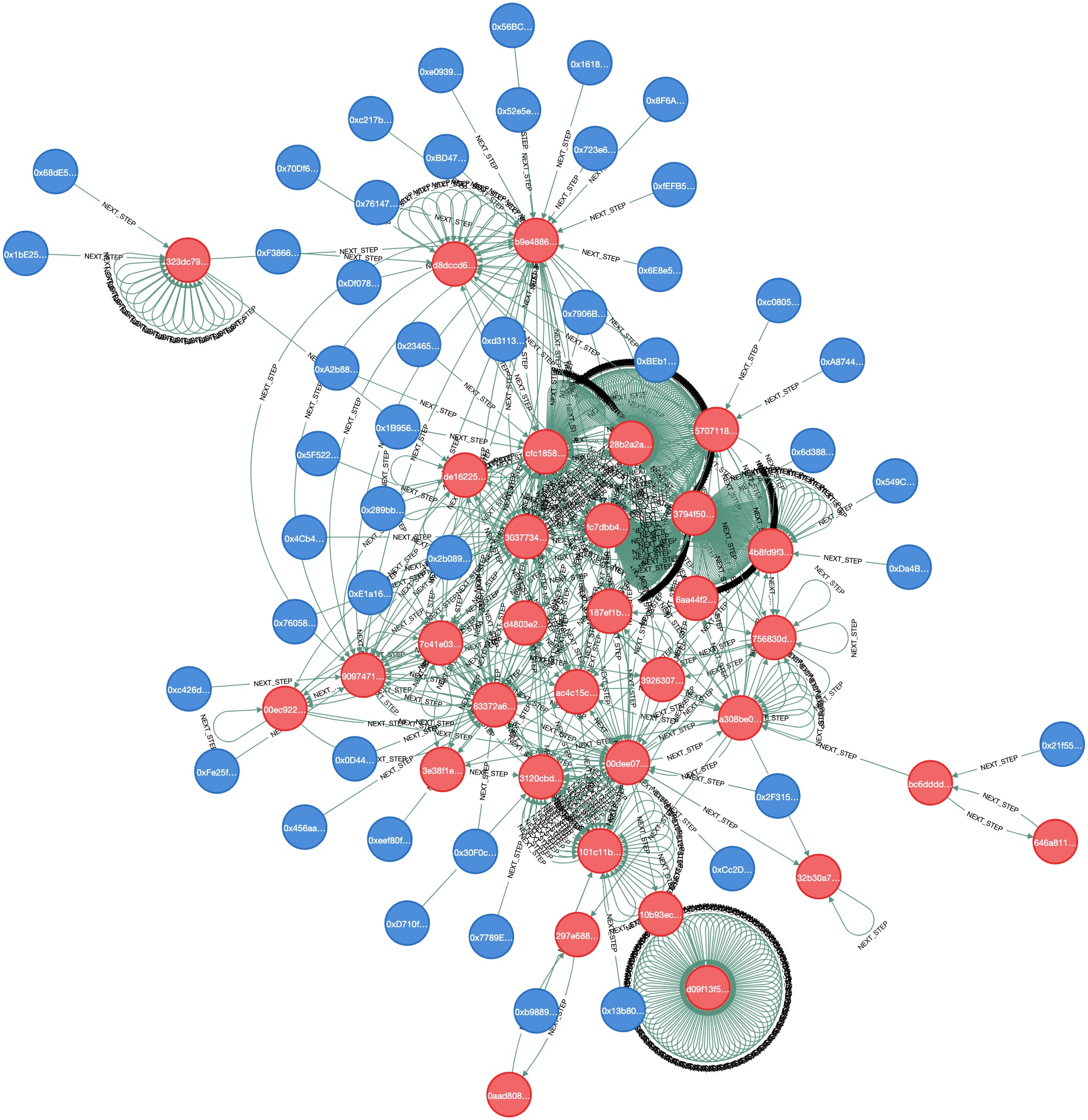}
        \caption{Cluster $C_4$}
    \end{subfigure}
    \begin{subfigure}[b]{0.3\textwidth}
        \centering
        \includegraphics[width=\textwidth]{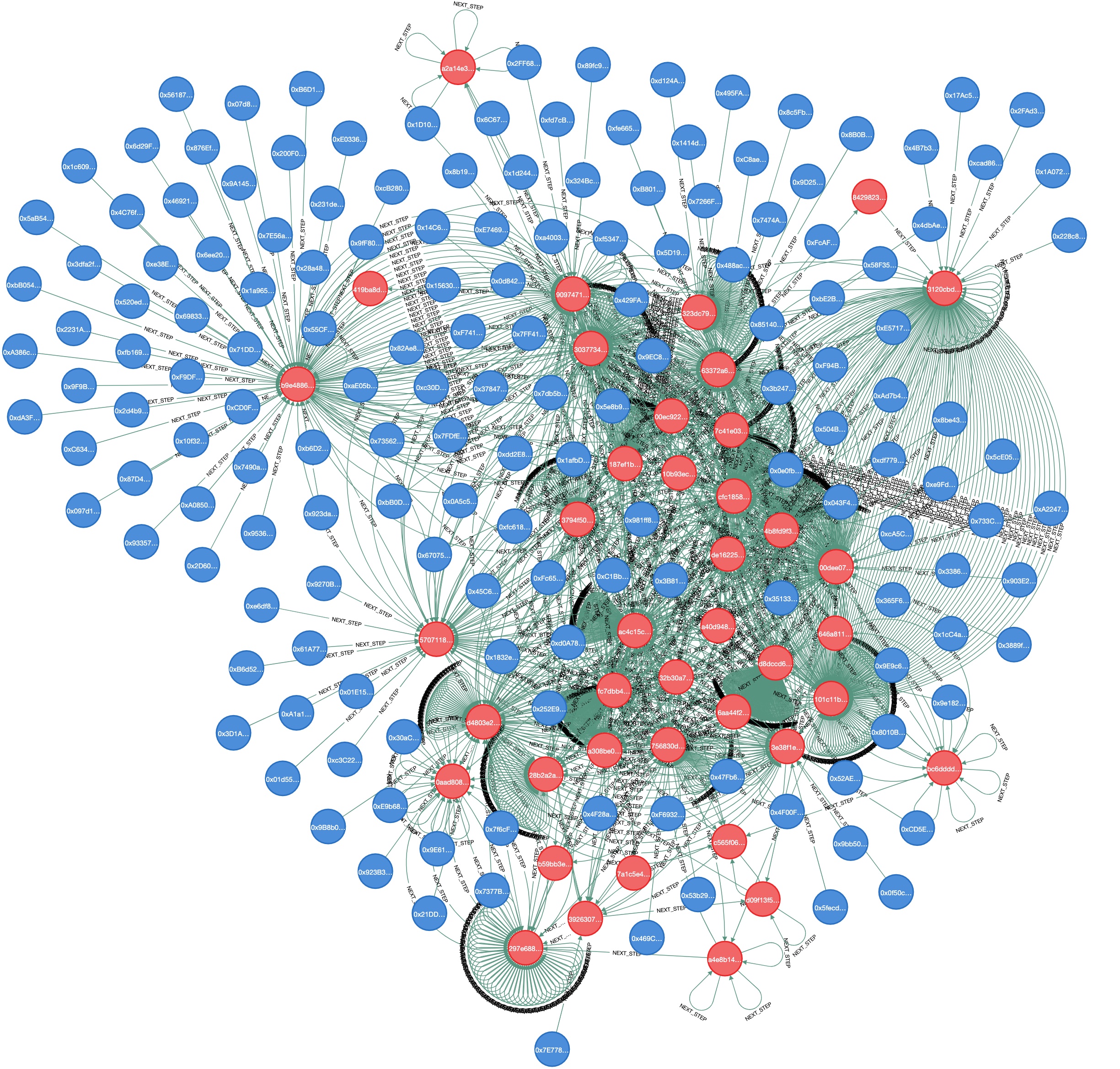}
        \caption{Cluster $C_5$}
    \end{subfigure}
    \begin{subfigure}{0.3\textwidth}
        \centering
        \begin{subfigure}{0.85\textwidth}
            \centering
            \begin{overpic}[width=\textwidth]{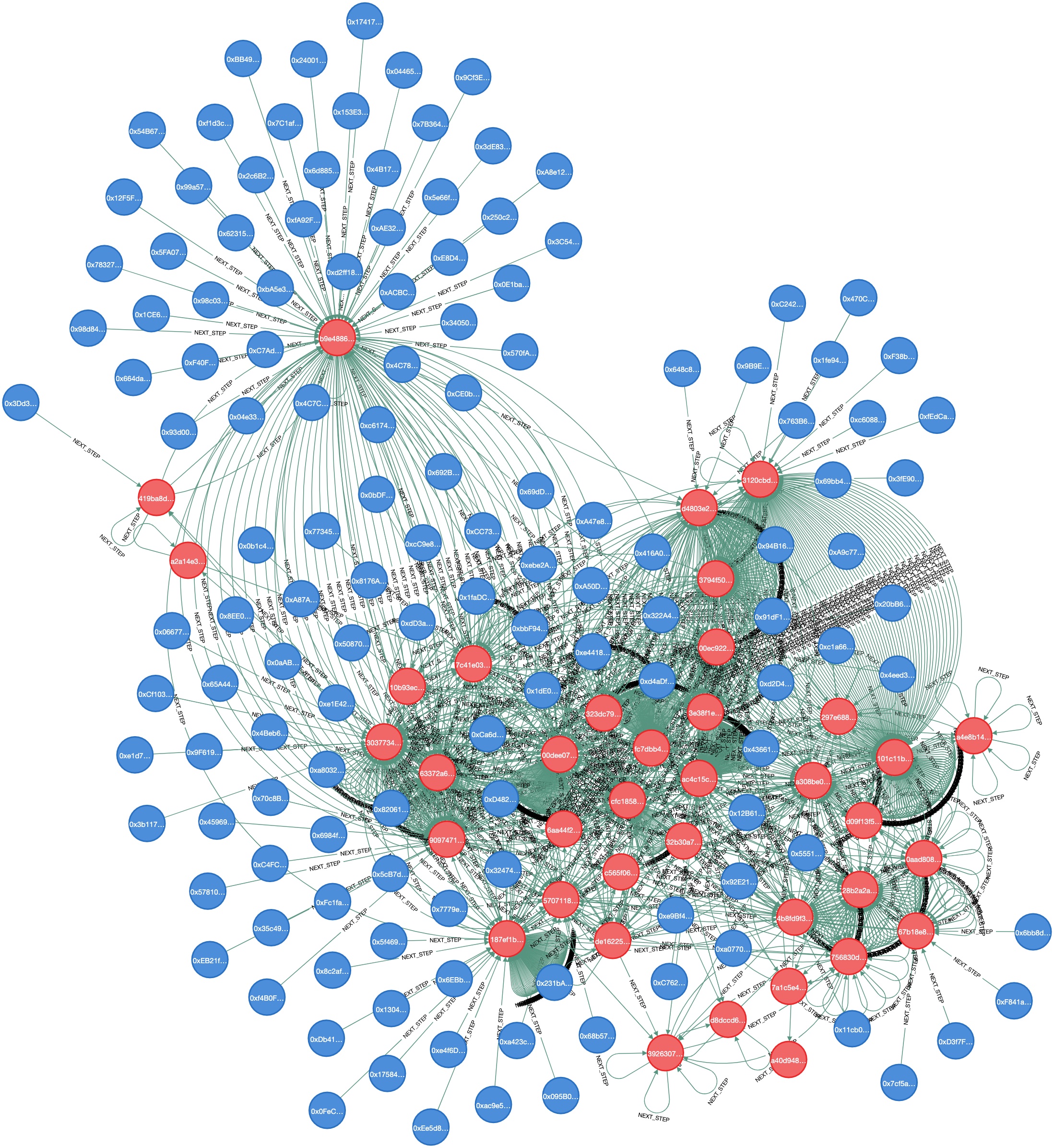} 
                \put(72,-20){ 
                    \begin{tikzpicture}
                        \filldraw[fill={rgb,255:red,255; green,0; blue,0}, draw=none] (1, 1.6) circle (0.1cm);
                        \node[right] at (1.3, 1.6) {Actions};
                        \filldraw[fill={rgb,255:red,0; green,125; blue,255}, draw=none] (1, 1) circle (0.1cm);
                        \node[right] at (1.3, 1) {Users};
                    \end{tikzpicture}
                }
            \end{overpic}
            \caption{Cluster $C_6$}
        \end{subfigure}
    \end{subfigure}
    \caption{Behavioural clusters from \textit{k}-means.}
    \label{fig:clusters-kmeans}
\end{figure*}
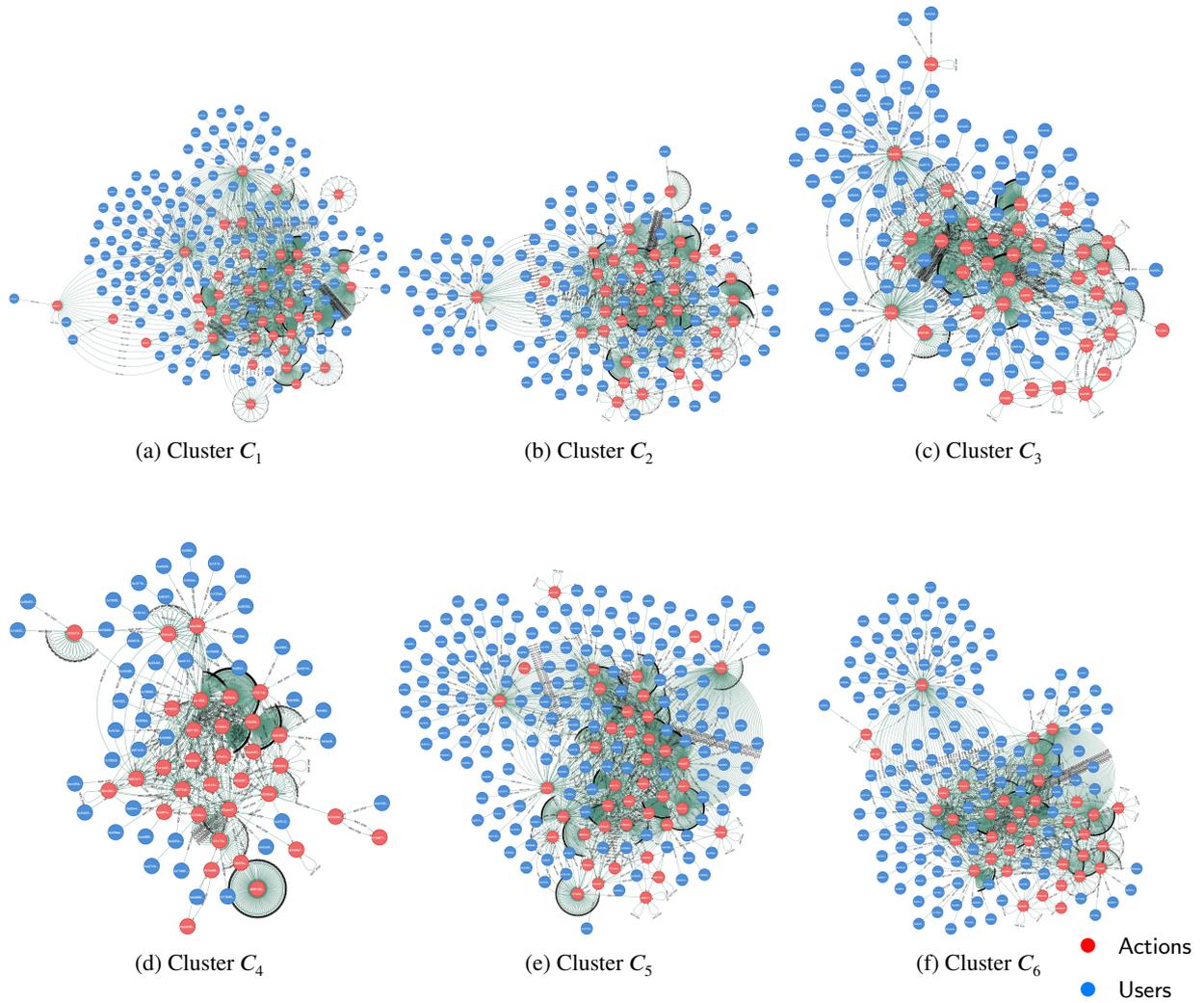
\section{Results and discussion}
\label{sec:results}
In this section, the results of the clustering process are presented. At first, a comparison of the clustering algorithms is provided. Based on that, one of the clustering methods was chosen, and behavioural clusters that were produced by this algorithm were presented. For each cluster, descriptions were provided based on multiple aspects and some visualisation examples are also presented.

\subsection{Comparison of the clustering algorithms}
Multiple clustering metrics were leveraged: the Silhouette Coefficient (SC) \cite{ROUSSEEUW198753}, the Davies-Bouldin Index (DBI) \cite{4766909} and the Calinski-Harabasz Index (CHI) \cite{doi:10.1080/03610927408827101}, to evaluate the performance of the various clustering algorithms. SC gives a value between -1 and 1 and the values closer to 1 present better clustering results. DBI ranges between 0 and $\infty$ and shows good clustering results when the value is close to 0. In CHI there is no specific range but higher values mean good clustering and lower values present poor results. The results are presented in Table~\ref{tab:clustering-metrics}. They show that \textit{k}-means performed the best based on SC and CHI and BIRCH outperformed all based on DBI. However, agglomerative clustering and bisecting \textit{k}-means produced very similar values as well which present a good performance so the usage of any of those is recommended. On the other hand, spectral clustering and affinity propagation presented poor results. We chose \textit{k}-means to present clustering results but any of the recommended clustering algorithms can be utilised in related research.

\subsection{Behavioural clusters}
In order to describe the different behavioural clusters that were constructed by \textit{k}-means, certain information was extracted as a base for their comparison. This can be seen in Table~\ref{tab:kmeans-clusters}. Multiple features are presented for all 6 clusters, they are named as follows in left to right order: the number of users in the cluster, the mean number of assets, tickets and packs that were used in any action, the length of the general user flow, the number of unique NFTs that the corresponding users utilised, the average time (in hours) the users spent in the game during the examined time period, two of our introduced metrics, asset usage (\textbf{\( \rho \)}) and NFT distribution (\textbf{\( \Phi \)}) that we introduce later and the action sequences for the general user flows. Those present how the actions follow each other in the corresponding general user flows. In order to show visualisation results, the clustering algorithm results were added as edge properties, which enables us to filter based on the clustering algorithm and the cluster number. The clusters generated by \textit{k}-means are visualised in Figure~\ref{fig:clusters-kmeans}. This presents how all user flow subgraphs of users of the clusters are merged together to a new subgraph of \( \mathbf{G} \).

\begin{table*}[ht!]
    \caption{Comparison of \textit{k}-means clusters.}
    \label{tab:kmeans-clusters}
    \centering
    \footnotesize
    \begin{NiceTabular}[hvlines]{ m[c]{1.5mm} m[c]{6.2mm} m[c]{8.7mm}m[c]{8.7mm}m[c]{8.7mm}m[c]{7mm}m[c]{6mm}m[c]{6.8mm} m[c]{7mm} m[c]{6.2mm} m[l]{52mm} } 
    \hline
     \textbf{\#} & \textbf{Users} & \textbf{Mean\# asset} & \textbf{Mean\# ticket} & \textbf{Mean\# pack} & \textbf{Flow length} & \textbf{NFTs} & \textbf{Time (hr)} & \textbf{Usage (\( \rho \))} & \textbf{Distr. (\( \Phi \))} & \multicolumn{1}{c|}{\textbf{Typical action sequence}}\\ \hline
     1 & 162 & 1.5578 & 0.0491 & 0.1639 & 11 & 441 & 52.19 & 0.50 & 2      & $a_0$, $a_1$, $a_2$, $a_3$, $a_0$, $a_4$, $a_5$, $a_6$, $a_7$, $a_8$, $a_9$
          \\ \hline
2 & 111 & 1.6725 & 0.0508 & 0.1525 & 11 & 574  & 64.53 & 0.40 & 2      & $a_0$, $a_0$, $a_5$, $a_7$, $a_{10}$, $a_{11}$, $a_{12}$, $a_{13}$, $a_6$, $a_{14}$, $a_9$
              \\ \hline
3 & 100 & 1.4878 & 0.0508 & 0.1525 & 11 & 188 & 69.62 & 0.54 & 2      & $a_0$, $a_1$, $a_5$, $a_{15}$, $a_7$, $a_9$, $a_{16}$, $a_{17}$, $a_{12}$, $a_6$, $a_{18}$
 \\ \hline
4 & 46 & 1.6011 & 0.0625 & 0.1875 & 15 & 820 & 247.20 & 0.50 & 3 & $a_0$, $a_{19}$, $a_0$, $a_{15}$, $a_{20}$, $a_{10}$, $a_{21}$, $a_{16}$, $a_{22}$, $a_6$, $a_{23}$, $a_{18}$, $a_{24}$, $a_{25}$, $a_{26}$
                   \\ \hline
5 & 161 & 1.2900 & 0.0769 & 0.1230 & 9 & 282 & 41.61 & 0.33 & 1 & $a_0$, $a_{27}$, $a_5$, $a_{15}$, $a_{17}$, $a_{10}$, $a_{16}$, $a_7$, $a_3$
                   \\ \hline
6 & 136 & 1.2849 & 0.0517 & 0.1034 & 9 & 280 & 69.36 & 0.62 & 1 & $a_0$, $a_{16}$, $a_5$, $a_2$, $a_{10}$, $a_{16}$, $a_6$, $a_{22}$, $a_{28}$
                   \\ \hline
    \end{NiceTabular}
\end{table*} 

For better visibility, a general user flow was also generated for each cluster. For every possible step (based on the average amount of steps the included users made within the examined period) in that cluster, a check was made for which action was the mode action to determine which one was the most frequently performed by the users of that cluster at the specific step, and that was added as the action for that specific step. In order to visualise this, a virtual user node was added as the starting user for all general user flows and the edges in between nodes were added as a new type of edge called VIRTUAL\_STEP which has the \textit{k}-means cluster number as property. In Figure~\ref{fig:general-user-flows}, the general user flows for the behavioural clusters are presented. However, these graphs also include all the NFT assets that were used by the actions of the general user flows. In the following, all produced clusters are described, which we summarise in Table~\ref{tab:cluster-descriptions}. In both Table~\ref{tab:kmeans-clusters} and Table~\ref{tab:cluster-descriptions}, \# refers to the cluster number.

\begin{table}[]
\caption{Summary of cluster descriptions.}
\label{tab:cluster-descriptions}
\resizebox{0.48\textwidth}{!}{%
\begin{tabular}{|l|l|l|l|}
\hline
\multicolumn{1}{|c|}{\textbf{\#}} & \multicolumn{1}{c|}{\textbf{Active}} & \multicolumn{1}{c|}{\textbf{Asset usage}} & \multicolumn{1}{c|}{\textbf{Time}} \\ \hline
1 & No & High, not distributed & Low  \\ \hline
2 & No      & Low, not distributed     & Medium  \\ \hline
3 & Semi-active          & High, not distributed & Medium \\ \hline
4 & Yes         & High, distributed     & Long   \\ \hline
5 & No         & Low, not distributed     & Short   \\ \hline
6 & No         & High, not distributed     & Medium   \\ \hline
\end{tabular}
}
\end{table}

In the summary Table~\ref{tab:cluster-descriptions}, the asset usage is calculated by two metrics: \textbf{\( \rho \)} and \textbf{\( \Phi \)}. \textbf{\( \rho \)} refers to high/low asset usage by calculating the ratio of the actions that have any type of asset usage as shown in Eq~\ref{eq:use}, while \textbf{\( \Phi \)} presents whether these NFT assets are well distributed among the actions by checking the number of nodes that have the majority of the USED\_BY relations. This is detailed in Eq~\ref{eq:distr}. In both cases, if a node has multiple USED\_BY relations to a single NFT, we only counted one relation. Asset usage is high if \( \mathbf{\rho} \geqq \alpha \) where \( \alpha \) is the usability factor which determines the ratio of the action nodes that must have NFTs associated with them. NFTs are well distributed if the number of nodes, that have the majority ratio of the USED\_BY relations where this ratio is determined by the distribution factor \( \beta \), are higher than the distribution number of nodes that are represented by \( \gamma \). They are calculated as follows:

\begin{equation}
\centering
\mathbf{\rho} = \frac{\sum_{i=1}^{m} x_i \cdot \mathcal{I}(x_i > 0)}{m}
\tag{1}
\label{eq:use}
\end{equation}
where \( \mathcal{I}(x_i > 0) \) is the indicator function that equals 1 if \( x_i > 0 \) and 0 otherwise, where \( x_i \) is the number of USED\_BY relations associated with a particular node \( \mathbf{n} \) and \( m \) is the number of nodes in the general user flow.

\begin{equation}
\centering
\begin{aligned}
\mathbf{\Phi} = \min \left\{ k \mid \sum_{i=1}^{k} x_i \geq \beta \cdot \sum_{i=1}^{m} x_i \right\}
\end{aligned}
\tag{2}
\label{eq:distr}
\end{equation}
where \( x_i \) represents the number of USED\_BY relations for each node, \( m \) is the total number of nodes in the general user flow and \( k \) is the minimum number of nodes needed to reach or exceed the set threshold.

Based on the examination of the general user flows visualised in Figure~\ref{fig:general-user-flows}, we chose 0.5 for \( \alpha \), 0.6 for \( \beta \) and 2 for \( \gamma \). In Table~\ref{tab:kmeans-clusters}, values for both metrics can be seen.

Table~\ref{tab:cluster-descriptions} also presents information regarding user activity. When determining whether users in a cluster were active in the game or not, the following was considered:

\begin{itemize}
    \item Distributed NFT usage which suggests that they have participated in multiple parts of the game and not just focused on one aspect.
    \item The time the users spent in the game and the amount of actions they performed within the examined period. 
    \item Whether there is any element of the game that influences the users' performed actions.
\end{itemize}

The behavioural clusters generated by \textit{k}-means are described as follows  in the order of the ascending engagement level of the included users:

\begin{figure*}[h!]
    \centering
    \begin{subfigure}{0.46\textwidth}
        \centering
        \begin{subfigure}{0.9\textwidth}
            \centering
            \includegraphics[width=\textwidth]{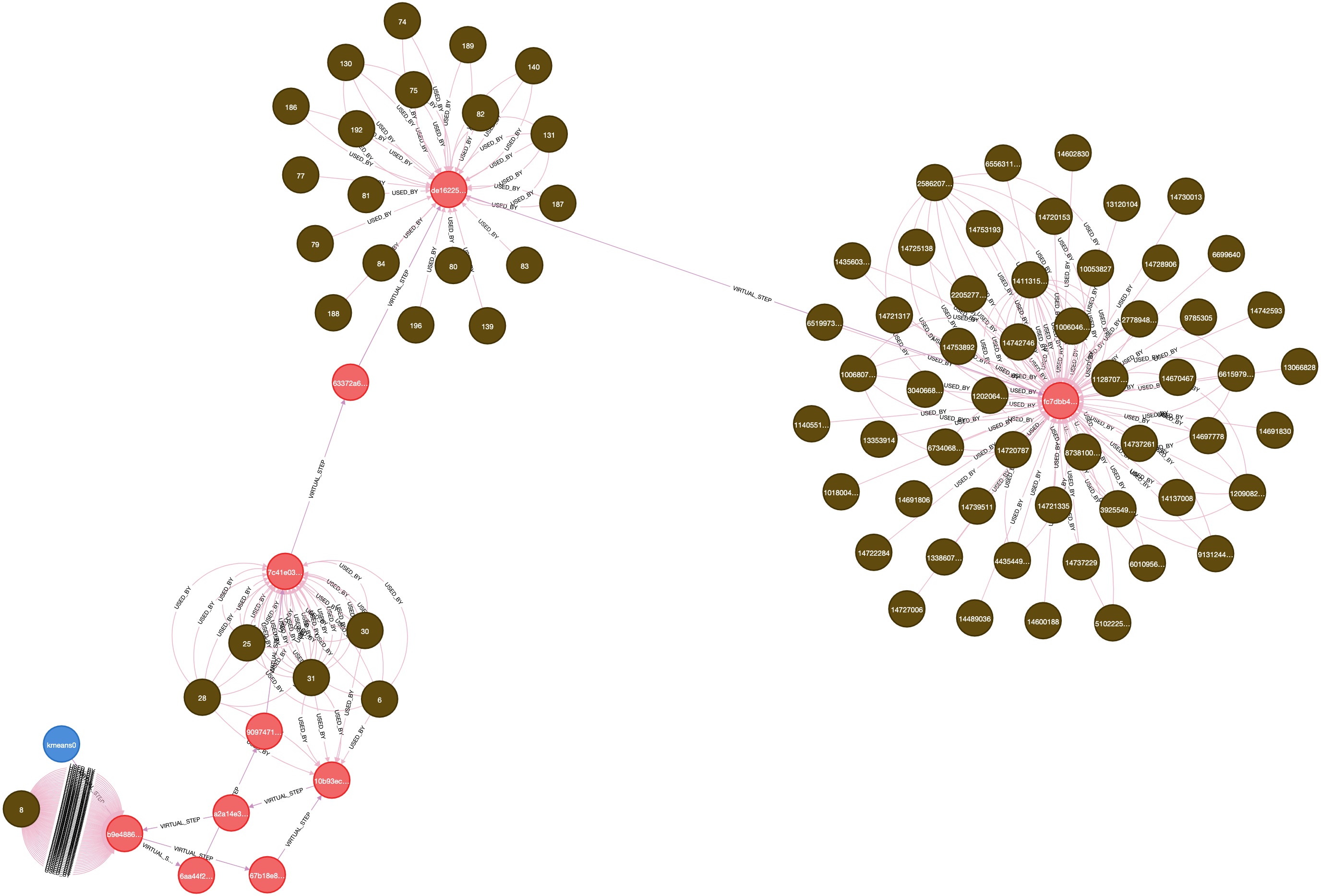}
        \end{subfigure}
        \caption{Cluster $C_1$ - Dropouts}
    \end{subfigure}
    \begin{subfigure}{0.46\textwidth}
        \centering
        \begin{subfigure}{0.85\textwidth}
            \centering
            \includegraphics[width=\textwidth]{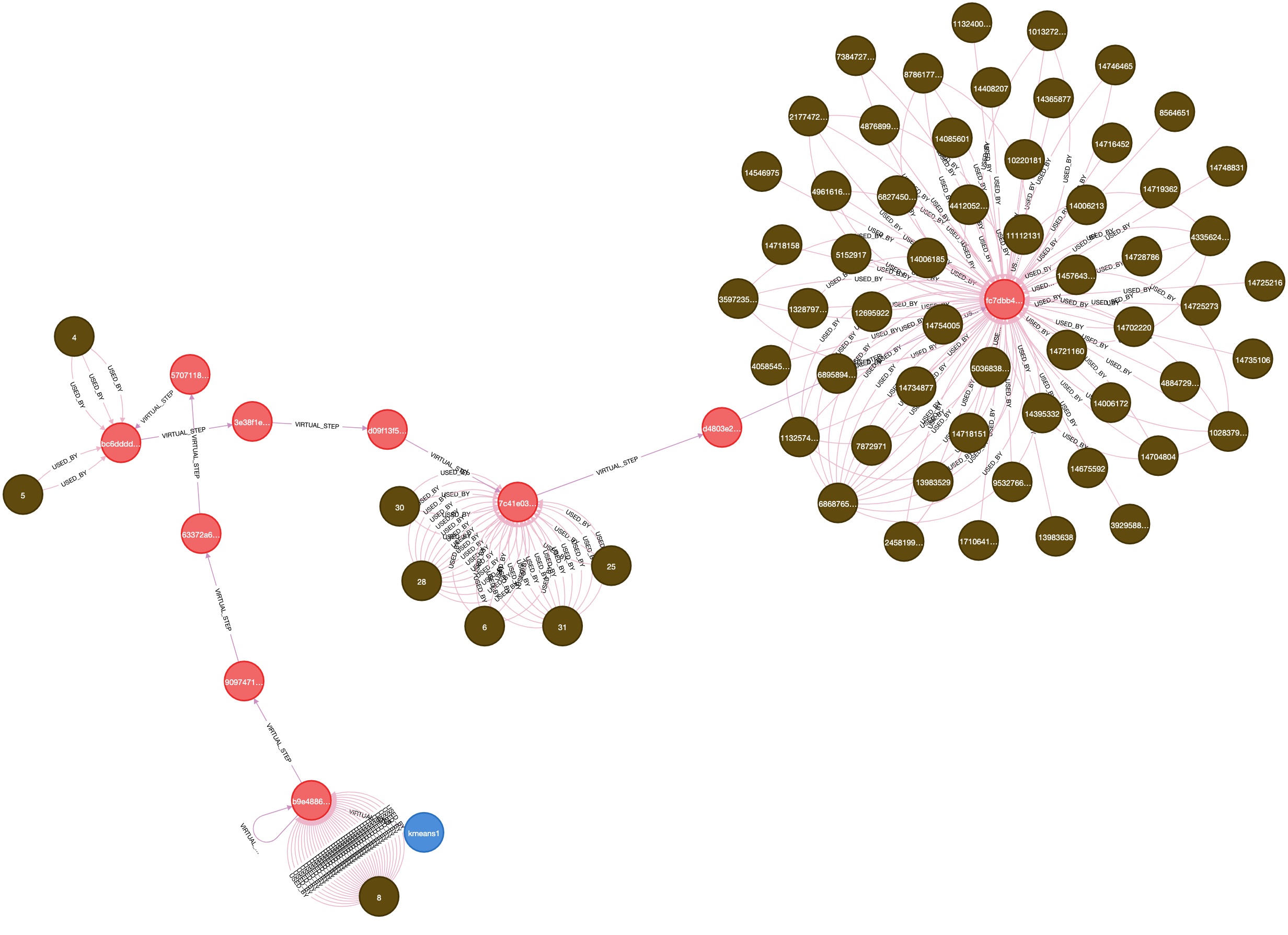}
        \end{subfigure}
        \caption{Cluster $C_2$ - Brief Engagers}
    \end{subfigure}

    \vspace{1cm}
    
    \begin{subfigure}{0.46\textwidth}
        \centering
        \begin{subfigure}{0.8\textwidth}
            \centering
            \includegraphics[width=\textwidth]{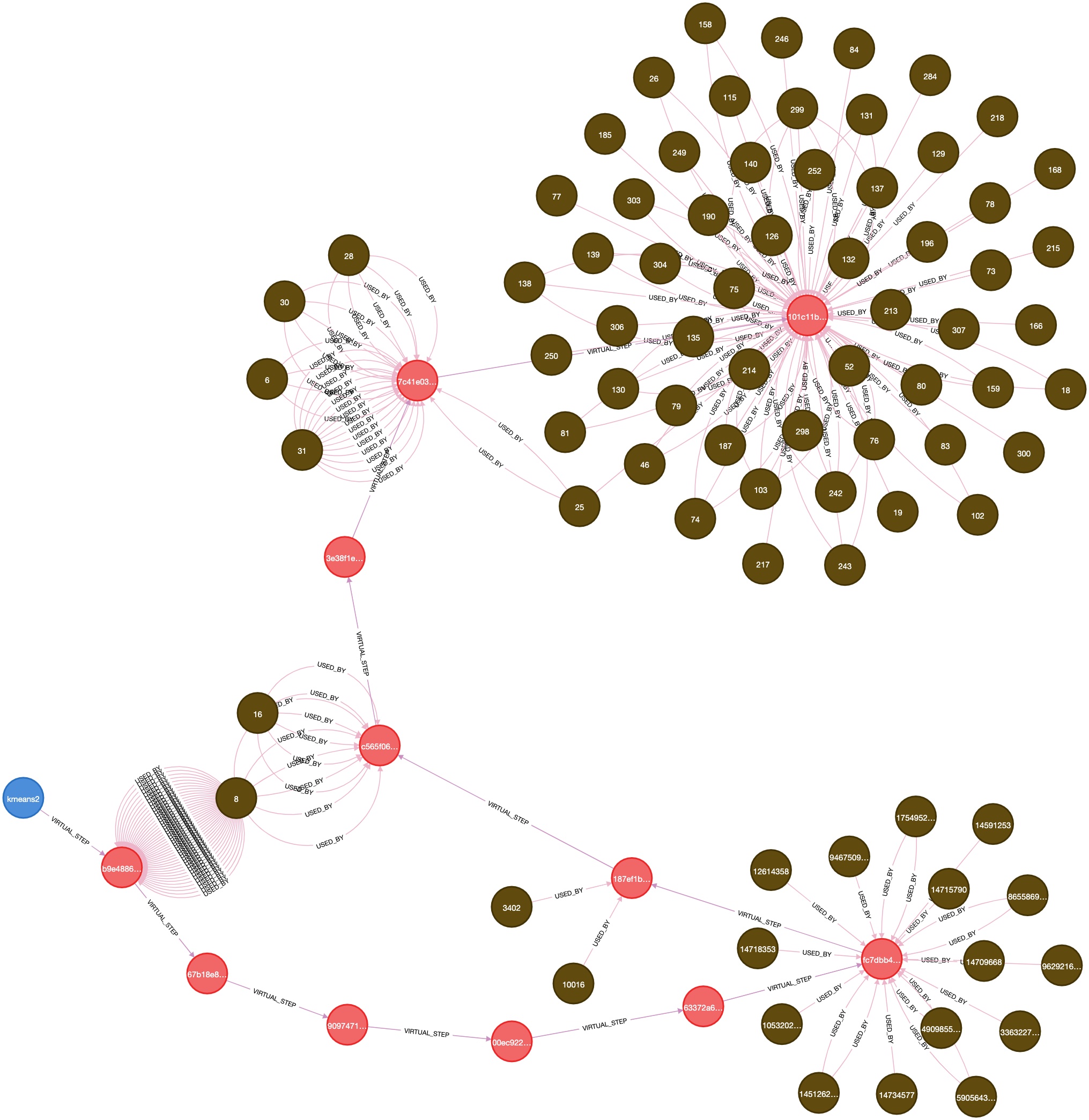}
        \end{subfigure}
        \caption{Cluster $C_3$ - Semi-active Users}
    \end{subfigure}
    \begin{subfigure}{0.46\textwidth}
        \centering
        \begin{subfigure}{0.85\textwidth}
            \centering
            \includegraphics[width=\textwidth]{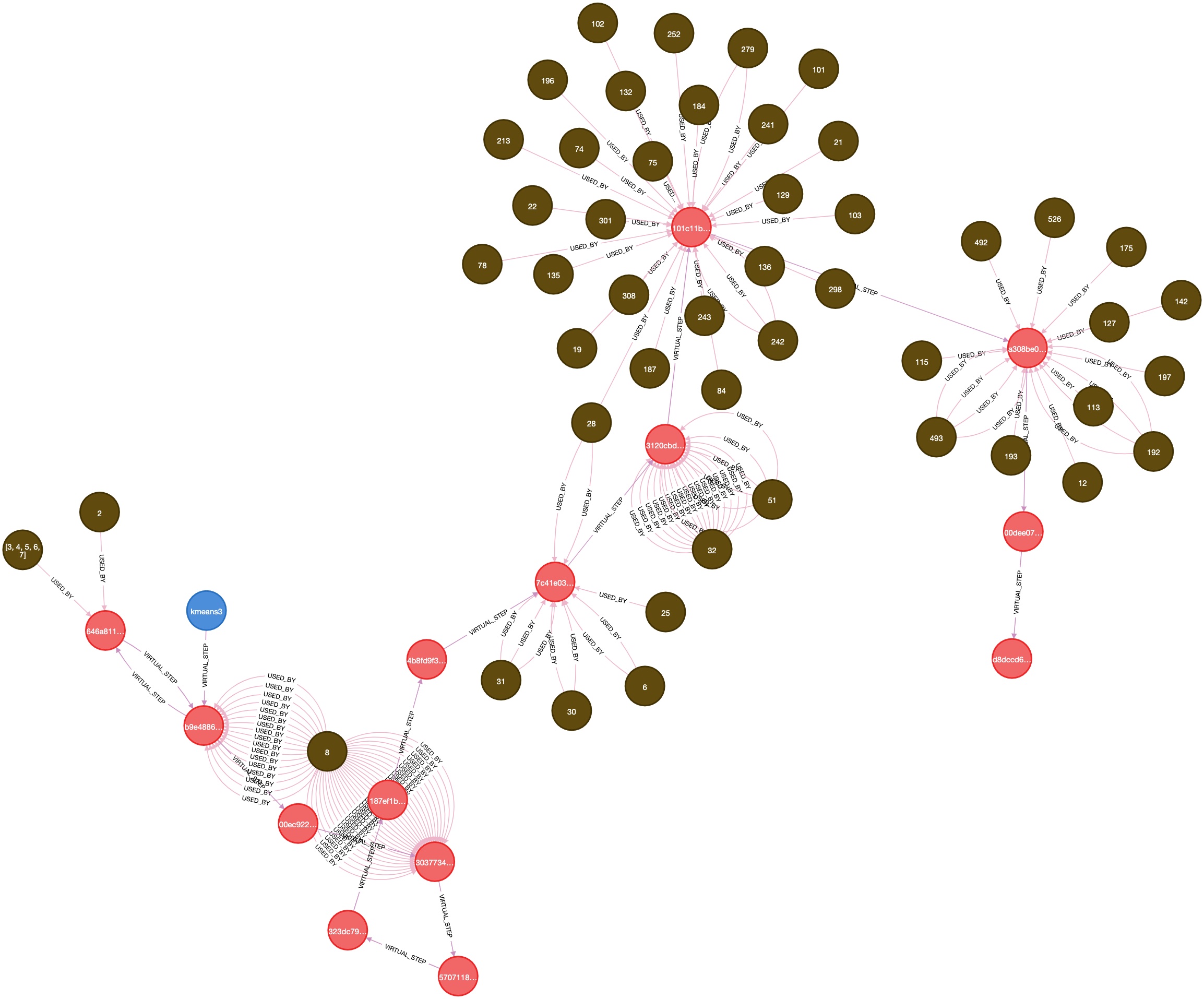}
        \end{subfigure}
        \vspace{0.65cm}
        \caption{Cluster $C_4$ - Active Users}
    \end{subfigure}

    \vspace{1cm}

    \begin{subfigure}{0.46\textwidth}
        \centering
        \begin{subfigure}{0.6\textwidth}
            \centering
            \includegraphics[width=\textwidth]{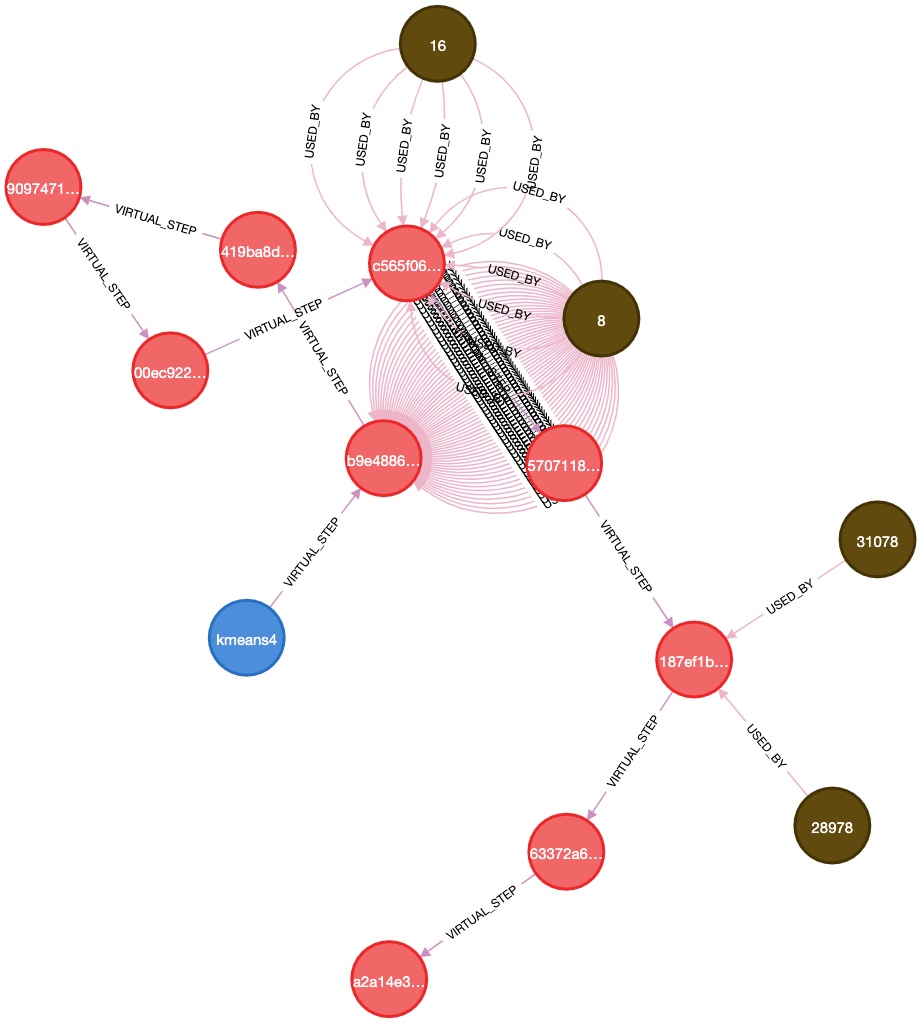}
        \end{subfigure}
        \caption{Cluster $C_5$ - Inactive Users with interest}
    \end{subfigure}
    \begin{subfigure}{0.46\textwidth}
        \centering
        \begin{subfigure}{0.85\textwidth}
            \centering
            \begin{overpic}[width=\textwidth]{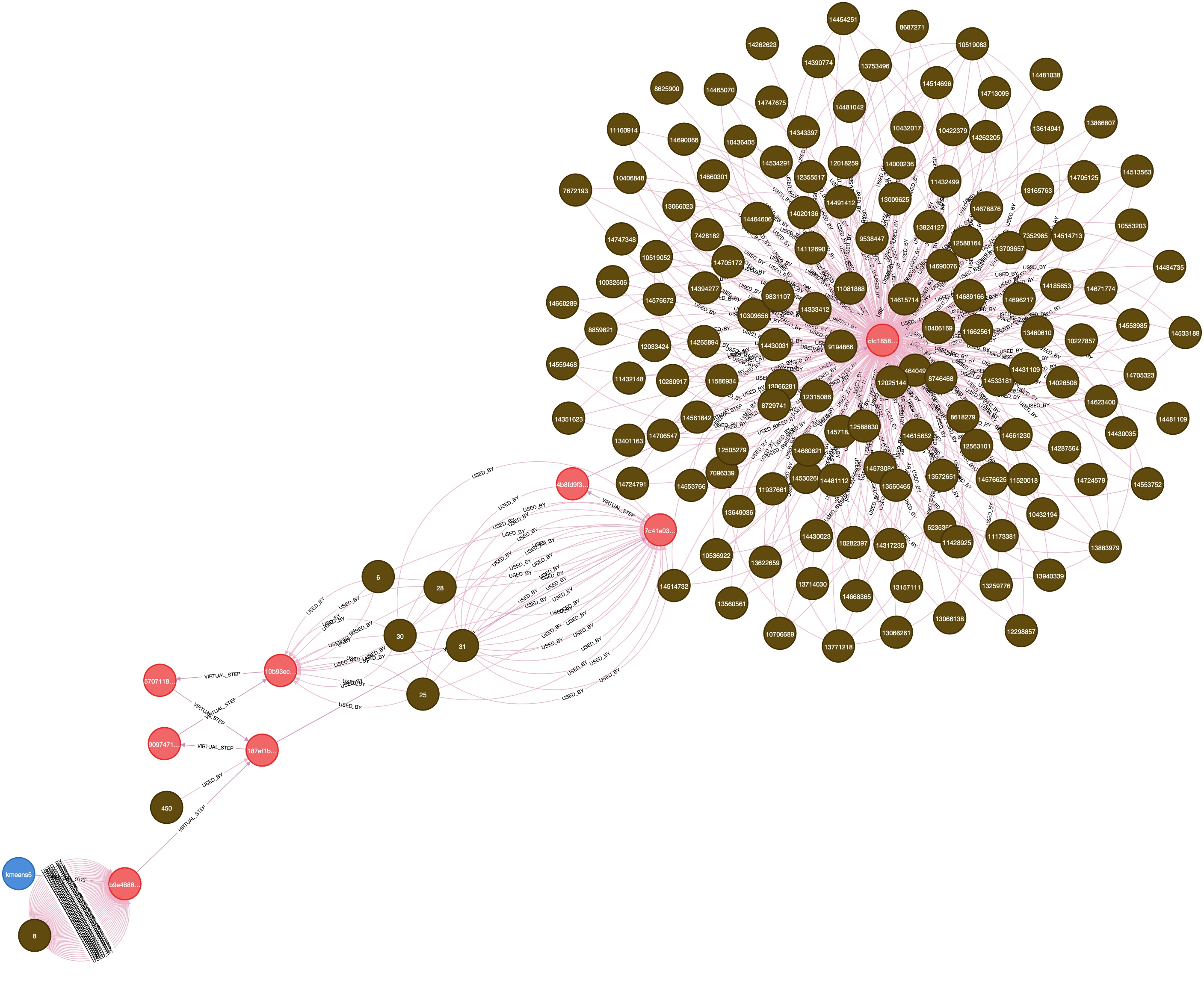}
                \put(75,0){
                    \begin{tikzpicture}
                        \filldraw[fill={rgb,255:red,255; green,0; blue,0}, draw=none] (1, 1.6) circle (0.1cm);
                        \node[right] at (1.3, 1.6) {Actions};
                        \filldraw[fill={rgb,255:red,0; green,125; blue,255}, draw=none] (1, 1) circle (0.1cm);
                        \node[right] at (1.3, 1) {Users};
                        \filldraw[fill={rgb,255:red,123; green,63; blue,0}, draw=none] (1, 0.4) circle (0.1cm);
                        \node[right] at (1.3, 0.4) {NFTs};
                    \end{tikzpicture}
                }
            \end{overpic}
            \caption{Cluster $C_6$ - Inactive Users with no interest}
        \end{subfigure}
    \end{subfigure}
    \caption{Generated general user flows.}
    \label{fig:general-user-flows}
\end{figure*}

\paragraph{Cluster $C_5$ - Inactive Users with interest:}
The short general user flow visualised in Figure~\ref{fig:general-user-flows}e, the low asset usage, mean asset and pack values and the lowest associated time period shows that these users were highly inactive within the period. However, the mean ticket value is the highest among all clusters which means that lottery-related activities may interest them and can be a key element in increasing potential engagement. As a result of this the users who belong to this cluster can be called inactive users with interested.

\paragraph{Cluster $C_6$ - Inactive Users with no interest:}
Similar behaviour to Cluster $C_5$ but with an increased amount of time spent in the application. The overall asset usage is higher but it is extremely low in all actions but one that has the majority of the involved assets as presented in Figure~\ref{fig:general-user-flows}f. There is no clear indication what could increase the users' activity level, therefore, further behavioural analysis can be crucial for the developers to gain insights on the users' potential interests and habits that can be utilised to make changes in the application that can be beneficial for them. As there is no defined interest, users of this cluster are named inactive users with no interest.

\paragraph{Cluster $C_2$ - Brief Engagers:}
Users in this cluster can be considered not active because although they performed multiple types of actions within a time period that is considered average, they did not use NFTs in a lot of actions. The involved assets are crucial parts of this application thus low usage presents low user participation. Figure~\ref{fig:general-user-flows}b presents this low asset usage. Application providers can also leverage analysis of this cluster to determine causes of low engagement after trial. The included users can be called brief engages as they try the application but do not get involved in all parts.

\paragraph{Cluster $C_1$ - Dropouts:}
This cluster presents a common behaviour where the included users conducted various types of activities within a short time with high asset usage but they did not get invested in the application. Figure~\ref{fig:general-user-flows}a presents a not well distributed asset usage as the majority of the assets are centered around only two action nodes. NFT distribution is important as the application consists of multiple activities offered by in-game corporations and some of those involve NFTs and others are being performed without the need for any asset usage. If the NFTs are well distributed among actions, it can be assumed that the corresponding users are invested in multiple parts of the application. Analysis of this behavioural cluster can provide information on why certain users did not get engaged in the application after conducting various types of activities in it. These users are called dropouts.

\paragraph{Cluster $C_3$ - Semi-active Users:}
Users of this cluster performed multiple actions with high asset usage within an average time period but with a poor distribution as presented in Figure~\ref{fig:general-user-flows}c. They can be considered semi-active users as they engaged in the game but the distribution presents that they did not get fully involved even though they spent more time in the application than users of Cluster $C_2$. Analysis of this cluster can help developers to adopt strategies that can potentially increase the corresponding users' engagement.

\paragraph{Cluster $C_4$ - Active Users:}
This cluster groups the users that presented active user behaviour within the examined period. Figure~\ref{fig:general-user-flows}d presents the longest general user flow that shows high asset usage with good distribution. The values for mean asset, ticket and pack also present that they utilised the game assets when participating in multiple types of in-game activities. This cluster has the longest time period which means that these users not only engaged but presented continuos active behaviour. However, this cluster has the lowest associated user base thereby, it can be stated that active user behaviour is rare in this dApp. Behavioural analysis of the involved users can enable the establishment of engagement models of active user behaviour which can be utilised by the game providers to check which parts of the application boost active engagement.

\begin{figure}[ht!]
\centerline{\includegraphics[width=0.45\textwidth]{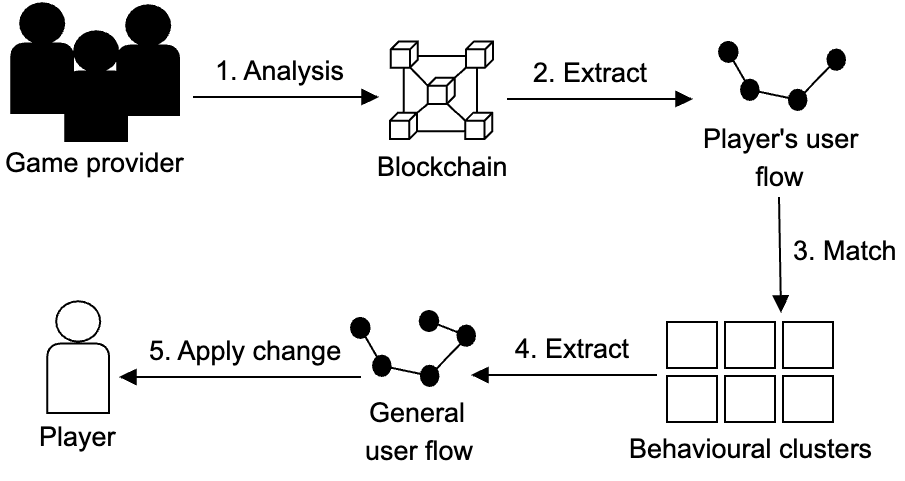}}
\caption{Privacy threat model.}
\label{fig:threat_model}
\end{figure}

\subsection{Privacy threat model}
The modelled version of a potential privacy threat is presented in this sub-section, which can be seen in Figure~\ref{fig:threat_model}. Game providers can conduct similar types of blockchain analysis of their games, establishing behavioural clusters. If a game provider would like to deduce what to add or remove in the game in order to increase the engagement of a certain player, the provider can extract the player's corresponding user flow. Then that can be matched to the behavioural clusters' general user flows. From the cluster that matches, the game provider can extract the general user flow, which shows the usual activities of the belonging users and other in-game behavioural information in regard to how the corresponding users usually behave. The provider then can form and apply a change based on that which specifically targets the specific player. In the following, a couple of affected areas are presented:

\paragraph{GameFi:} This threat model presents that this type of research has a use case in the GameFi sector. This is further supported by the examples given in the previous sub-section. For example, behavioural information from an active cluster can be utilised to reveal which activities boost interest and information extracted from lesser active clusters can be leveraged to conduct churn rate-related analysis. Results from these analyses can enable game providers to design more effective engagement and incentive models.

\paragraph{Secondary NFT marketplace:} Users can buy assets in the primary NFT marketplace that is associated with the particular dApp. However, because of full ownership of assets on the blockchain, these users can sell these assets on other platforms that are not related to the original dApp as well. We refer to these as secondary NFT marketplaces. Analysis of the primary marketplace, secondary marketplace or both can be also beneficial if a malicious actor would like to learn more about a certain user's behaviour. Therefore, the pipeline can be utilised in this area as well. Data from the smart contracts that are associated with the marketplaces can be collected and behavioural clusters can be established based on it. Targeted users' user flows can then be matched to the general user flows of the clusters and information from the matched cluster can be utilised against the targeted user. For example, it can be learned whether he/she is willing to participate in an auction or not.

\paragraph{Metaverse:} In a metaverse setting malicious attackers can utilise similar techniques to construct social profiles of users based on their activities and interactions. They can extract the users' user flows and match them to general user flows thereby, gaining knowledge of common activities and habits that are present in the matching cluster. These profiles can be leveraged to create fake avatars, which can be utilised to impersonate the original users or perform scams to a targeted user base (users within a certain cluster), similarly to how they are presented in \cite{9880528}. To present an example, they can check which cluster includes the most active users and target them with scams, as they are more likely to respond because of their higher participation rate. As a result of the possibility of establishing behavioural clusters or user profiles, blockchain applications in privacy-sensitive areas such as healthcare or governance could face similar problems.

\section{Conclusion}
\label{sec:conclusion}
Public blockchains have the potential to be used in multiple application areas. However, the traceability of transaction data by anyone presents a privacy challenge as blockchain data analysis can potentially be used to extract user behaviour. This work presented a process to extract and define user behaviour clusters based on blockchain transaction data that was collected from a chosen dApp which is a blockchain-based game. From the data collection, unique actions were identified and leveraged to present user action flows that described the user's activities and asset usage. These user flows were utilised as input to a GNN model to provide embeddings, which were then used in clustering algorithms to put them into diverse clusters. These clusters can be visualised and analysed to discuss the related user behaviour. The first resulting behavioural clusters present multiple information regarding the participating users like their activity level and asset usage.

In future work, the time period provided for the user flows will be increased so long-term user behaviour can also be analysed. This means a wider time period that corresponds to the current timestamps and also the examination of user behaviour from a time period that is very distant from the current timestamps. We assume that the analysis between multiple periods may provide more expressive results. Currently, we only consider data from game-related smart contracts, but combining that with transaction information from secondary NFT markets and market analysis may extend information presented in the form of behavioural clusters. Data from other blockchain-based applications DeFi services will be collected and it will be investigated whether there are general behavioural patterns across multiple dApps.












\bibliographystyle{elsarticle-num}

\bibliography{cas-refs}

\end{document}